\documentclass[runningheads]{llncs}
\usepackage{graphicx}

\usepackage{tikz}
\usepackage{comment}
\usepackage{amsmath,amssymb} 
\usepackage{color}
\usepackage{multirow}
\usepackage{gensymb}
\usepackage{makecell}

\usepackage[noend]{algpseudocode}
\usepackage{algorithmicx,algorithm}

\begin{document}
\pagestyle{headings}
\mainmatter
\def\ECCVSubNumber{1701}  

\title{ASpanFormer: Detector-Free Image Matching with Adaptive Span Transformer} 

\titlerunning{ASpanFormer}
%
\author{Hongkai Chen \and
Zixin Luo \and
Lei Zhou\and
Yurun Tian \and
Mingmin Zhen \and\\
Tian Fang \and
David McKinnon \and
Yanghai Tsin \and
Long Quan}
\authorrunning{H. Chen et al.}
%
\institute{HKUST ~~~~~~~~~~~~~~ Apple Inc. }
\maketitle

\begin{abstract}

Generating robust and reliable correspondences across images is a fundamental task for a diversity of applications. To capture context at both global and local granularity, we propose ASpanFormer, a Transformer-based detector-free matcher that is built on hierarchical attention structure, adopting a novel attention operation which is capable of adjusting attention span in a self-adaptive manner. To achieve this goal, first, flow maps are regressed in each cross attention phase to locate the center of search region. Next, a sampling grid is generated around the center, whose size, instead of being empirically configured as fixed, is adaptively computed from a pixel uncertainty estimated along with the flow map. Finally, attention is computed across two images within derived regions, referred to as attention span. By these means, we are able to not only maintain long-range dependencies, but also enable fine-grained attention among pixels of high relevance that compensates essential locality and piece-wise smoothness in matching tasks. State-of-the-art accuracy on a wide range of evaluation benchmarks validates the strong matching capability of our method.

\keywords{Image Matching, Visual Localization, Pose Estimation, Transformer}
\end{abstract}

\begin{figure}[t]
	\centering
	\includegraphics[width=0.98\textwidth]{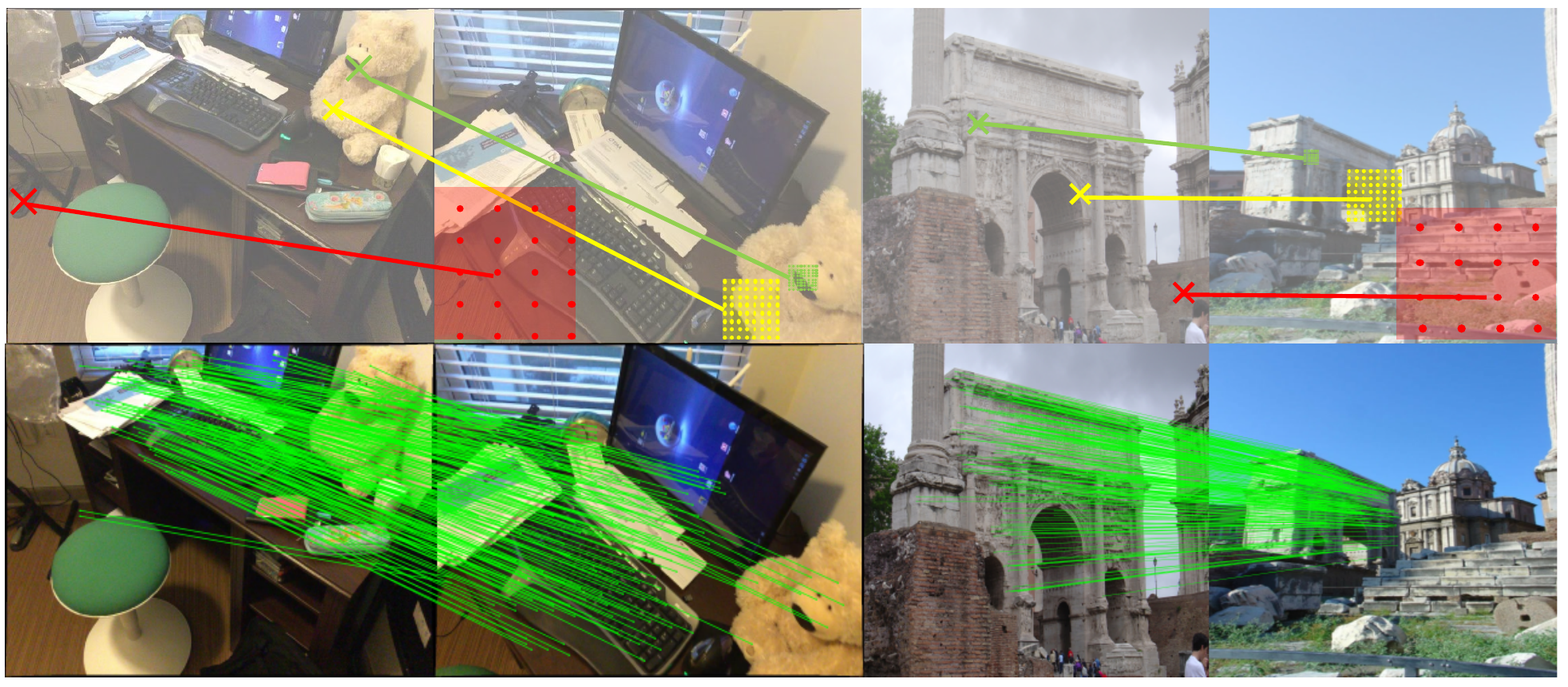}
	\caption{An illustration of the proposed adaptive attention span (top row) and final dense matching results (bottom row). Particularly, in the top row, a rectangle with $8\times8$ uniform sampling grid is drawn to explain the position and size of adaptive attention span. In addition, three typical types of correspondences are visualized. Easy match in \textcolor{green}{green} with rich texture, which can be well localized and matched with small local contexts. Hard match in \textcolor{yellow}{yellow} with little texture, which requires larger contexts to guide matching. Impossible match in \textcolor{red}{red} in non-overlapping or occluded region, which has a very large attention span to avoid falsely fitting to certain regions. With this design, we enable Transformer to adaptively capture necessary context according to matching difficulty.
	}
	\label{fig:matching}
\end{figure} 

\section{Introduction}

Image matching lays the foundation for various geometric computer vision tasks, including Structure from Motion (SfM)~\cite{schonberger2016structure,sfm2}, visual localization~\cite{sattler2012imageBMCV}, and Simultaneous Localization And Mapping (SLAM)~\cite{mur2015orb,orbslam2}. As a widely accepted pipeline for image matching, cross-image correspondences are usually established by matching a set of detected and described sparse keypoints, such as SIFT~\cite{lowe2004distinctive}, ORB~\cite{rublee2011orb}, or their learning-based counterparts~\cite{revaud2019r2d2,detone2018superpoint,luo2020aslfeat,d2net,contextdesc}. Despite its general effectiveness, this detector-based matching pipeline struggles in extreme situations, including large view point changes and textureless areas, due to the reliance on keypoint detector and context loss in feature description.

Concurrent with detector-based matching, another line of works focus on generating correspondences directly from raw images~\cite{sun2021loftr,jiang2021cotr,truong2021pdc,ncnet,li2020dual,truong2020glu,houghnet,ransasflow,tang2022quadtree}, where richer context can be leveraged while keypoint detection step can be eschewed. Earlier works~\cite{ncnet,rocco2020efficient,li2020dual} in detector-free matching often rely on iterative convolution upon correlation or cost volume to discover potential neighbourhood consensus. Recently, some works~\cite{sun2021loftr,jiang2021cotr,tang2022quadtree} base their methods on Transformer~\cite{vaswani2017attention,dosovitskiy2020image} backbone for better modeling of long-range dependencies. As a representative work, LoFTR utilizes self and cross attention blocks to update cross-view features, where Linear Transformer~\cite{katharopoulos2020transformers} is adopted to replace  global full attention in order to achieve manageable computation cost. Although proven effective, a concern about LoFTR is the lack of fine-level local interaction among pixel tokens, which could limit its capability to extract highly accurate and well-localized correspondences. This concern is deepened by the findings~\cite{tang2022quadtree} of Tang et al., which reveals that the cross attention map generated by LoFTR's Linear Transformer tends to diffuse among large areas instead of sharply focusing on actual corresponding regions.

To capture both global context and local details, we propose a Transformer-based detector-free matcher, equipped with a hierarchical attention framework. Our foundation processing blocks, referred to as Global-Local Attention (GLA) block, performs a coarse-level global attention at low resolution to acquire long-range dependencies, meanwhile, conducts fine-level local attention at high resolution within only a concentrated region around a current correspondence found through dense flow prediction.   

The key challenge for fine-level local attention is to determine the size of the attention span. A naive approach is to regard its size as a fixed hyper parameter, which, however, neglects the intrinsic matchability of different regions where the dependency of context varies. As shown in Figure~\ref{fig:matching}, regions in rich texture areas can be easily matched within a small neighbourhood, while the textureless areas are more uncertain about their correspondences and require larger context for matching, not to mention areas that lie out of overlapping regions and are impossible to be matched. To mitigate this issue, we introduce an adaptive attention span driven by probabilistic modelling, which can be adjusted for different locations based on underlying matching difficulty. We summarize our contributions in three aspects:

\begin{itemize}
  \item A hierarchical attention framework is proposed for feature matching, where attention operations are performed at different scales to enable both global context awareness and fine-grained matching.
  \item A novel uncertainty-driven scheme, based on probabilistic modelling of flow prediction, is proposed to adaptively adjust local attention span. Through this design, our network assigns varying size of contexts to different locations according to their essential matchability and context richness.
  \item State-of-the-arts results on extensive set of benchmarks are achieved. Our method outperforms both detector-free and detector-based matching baselines in two-view pose estimation. Further experiments on challenging visual localization also proves our method's potential to be integrated into c

  omplicated down-stream applications.  
\end{itemize}

\section{Related Works}

\subsection{Detector-Free Image Matching}
Differing from detector-based matching methods, which typical involve detecting~\cite{d2net,detone2018superpoint,luo2020aslfeat,revaud2019r2d2}, describing~\cite{hardnet,l2net,geodesc,contextdesc,caps} and matching~\cite{sarlin2020superglue,chen2021sgm,zhang2019oanet,pointcn,sun2020acne,adalam,Bian2020gms} a set of keypoints, detector-free matching consumes a pair of images and output correspondences in one shot. Thanks to the removal of keypoint detection stage, detector-free matching is able to capture richer contexts from original images, thus exhibits strong potential to match in extreme situations, such as low texture areas and repetitive patterns.

Despite the potential merits of detector-free matching, its popularity hardly outperforms detector-based methods during early deep learning times due to the intrinsic difficulties in robust and distinctive features. Recently, with the help of deep neural network, possibility is explored to build high performance detector-free matching frameworks based on deep features, which can roughly be classified into two categories: cost volume-based methods~\cite{ncnet,truong2020glu,truong2021pdc,rocco2020efficient,li2020dual,GOCor} and Transformer-based methods~\cite{sun2021loftr,jiang2021cotr,tang2022quadtree}. Both kinds of methods leverage strong signals in deep features' correlation, either in form of correlation layer or cross attention, to guide correspondence search and feature update. Our method follows works on Transformer-based methods and employs multilevel cross attention for mutual feature update, encoding two-view contexts into original features for both global and local consensus.

\subsection{Global-Local Structure}
Balancing receptive field and interaction granularity is a long-standing issue for both cost volume-based and Transformer-based matching. To ensure global receptive field, cost volume based methods are often designed to perform convolution on large global correlation volume, while Transformer-based methods need to conduct attention among all pixel tokens in image pairs. Due to the high cost of global interaction, the input features are usually downsampled into coarse resolution~\cite{li2020dual,truong2020glu,jiang2021cotr} or being projected into low rank~\cite{sun2021loftr}, which to some degree limits the networks' capability for fined grained feature update. 

Complementary to global interaction, some methods propose to perform local interaction only within a certain region instead of a global field, enabling to process fine level features given a limited computation budget. This practice is especially common in cost volume based methods and are referred to as local correlation layer~\cite{truong2020glu,flownet2,truong2021pdc,raft}, where the cost volumes/vectors are only constructed around neighbourhood of each correspondence estimation. Intuitively, the idea of complementary global-local interaction can also be introduced to Transformer-based matcher. In our method, a global-local attention block is proposed for message passing across images, ensuring both global receptive field and fine level feature processing. Specially, instead of fixing span for local attention, we design an adaptive mechanism to determine the size of area that each pixel should attend to.

\subsection{Flow Regression and Uncertainty Modeling}
Flow maps depicts correspondence coordinates, which can either be absolute or relative, for each location in an image. Predicting correspondence coordinates from an image pair has been intensively investigated by works in optical flow estimation~\cite{flownet,flownet2,yin2018geonet,raft} and general dense image matching~\cite{truong2020glu,truong2021pdc,GOCor}. In these works, the flow maps are regressed from structured correlation volumes which are implicitly position-aware. Recently, a Transformer-based method, COTR~\cite{jiang2021cotr}, proves that flows can also be retrieved from positional-embedded features after several turns of attention update. 

Naturally, the reliability of flow estimation in each location is not equal and predicting associated confidence scores is essential for many scenarios. As an elegant framework for uncertainty prediction, some works~\cite{truong2021pdc,zhou2020kfnet,prob1,prob2,prob3} propose to use probabilistic model to jointly explain both flows estimations and their confidence. Inspired by above works, we propose a network that regresses a flow map for each attention block to guide local attention region and adjust the attention span adaptively based on uncertainty prediction.

\section{Methodology}

\begin{figure*}[t]
	\centering
	\includegraphics[width=0.98\textwidth]{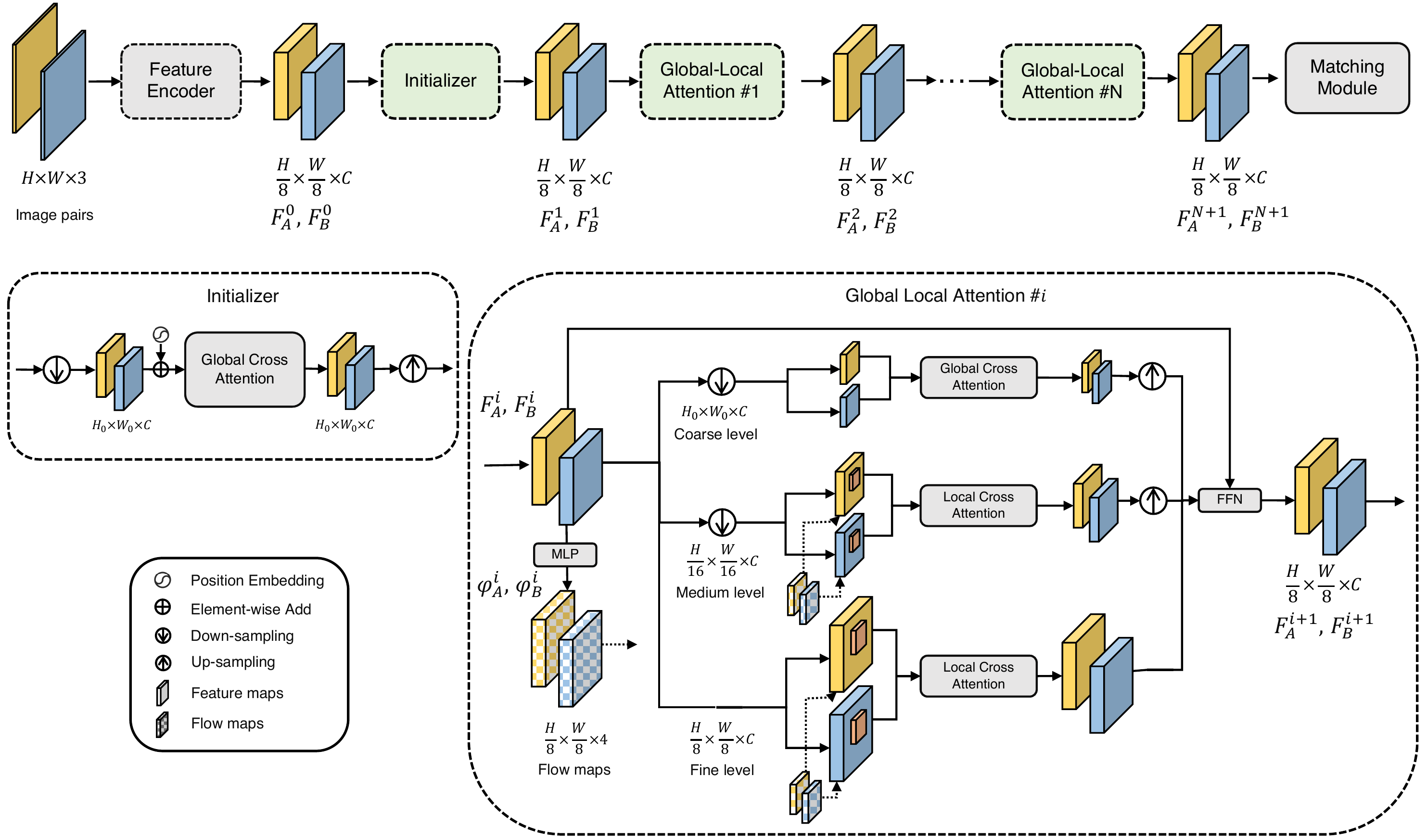}
	\caption{We use CNN backbone to extract initial features. After initialization, the features are fed into iterative GLA blocks for updating. A matching module is used to determine final matches.
	}
	\label{fig:net_arch}
\end{figure*}

We present an overview of our network structure in Figure~\ref{fig:net_arch}. Taking an image pair $I_A,I_B$ as input, our network produces reliable correspondences across images. The matching process starts with a CNN-based encoder to extract initial features $F_A^0,F_B^0$ for both images separately. After initialization, these features are turned into $F_A^1,F_B^1$ and fed into the proposed Adaptive Span Transformer (ASpanFormer) module for updating, which is composed of iterative global-local attention (\textbf{GLA}) blocks with hierarchical structure. Particularly, for each GLA block, we regress auxiliary flow maps $\phi_A,\phi_B$ describing correspondence coordinates (flows) and their uncertainty. Instead of adopting these flow maps as our correspondence output, we use them to guide local cross attention, enabling adaptive local attention span according to matching uncertainty. After $N$ GLA blocks, the updated features $F_A^{(N+1)},F_B^{(N+1)}$ are use to construct coarse level matches, which will be further refined into final correspondences.

In the following part, we demonstrate the details of each individual block as well as the underlying insights.

\subsection{Preliminary}
Before introducing the structure of our network, we first clarify necessary notations and concepts.  

\subsubsection{Attention.} As the key operation in Vision Transformer, attention is defined over a set of query ($Q$), key ($K$) and value ($V$) vectors as
\begin{align}
 M=\text{Att}(Q,K,V)=\text{softmax}(QK^T)V,
\end{align}
where $Q,K,V$ are linear projections of upstream features $F$ and $M$ is retrieved message. More specially, in the context of cross attention, $Q$ are derived from source features $F_s$ and $K,V$ vectors are derived from target features $F_t$. $M$ is used to update source features $F_s$ through a feed forward network (FFN), which involves concatenation, layer normalization and linear layers.
\begin{align}
 F_s^{i+1}=\textbf{FFN}(F_S^i,M).
\end{align}
Typically, in each pass, the position of source/target features can be switched and cross attention is performed symmetrically.

\subsubsection{Flow map.}
Flow maps $\phi_A,\phi_B \in R^{H\times W \times 2}$ depict the correspondence relationship between an image pair $I_A,I_B \in R^{H\times W}$, such that for any location $(i,j)$ in an image, $I_A[i,j]\leftrightarrow I_B[\phi_A[i,j]],  I_B[i,j]\leftrightarrow I_A[\phi_B[i,j]]$. Here, $\leftrightarrow$ denotes that the points on two sides are correspondences.

Instead of depicting simple correspondences, a stream of works~\cite{truong2021pdc,zhou2020kfnet,prob1,prob2,prob3} proposes to model flow field with a probabilistic framework. Following these works, we model the flow field as a Gaussian distribution defined by a set of parameters. More specifically, assuming conditional independence among pixels, two flow maps $\phi_A,\phi_B \in R^{H,W,4}$ are predicted, such that $\phi[i,j]=[u_x^{ij},u_y^{ij},\sigma_x^{ij},\sigma_y^{ij}]$, where $(u_x^{ij},u_y^{ij})$ are estimated correspondence coordinates and $(\sigma_x^{ij},\sigma_y^{ij})$ are standard deviations. The probability for $I_A[i,j]\leftrightarrow I_B[x,y]$ is given by

\begin{align}
P(x,y|\phi_A[i,j])= \frac{1}{2\pi\sigma_x^{ij}\sigma_y^{ij}}\text{exp}(-\frac{(x-u_x^{ij})^2}{2{\sigma_x^{ij}}^2}-\frac{(y-u_y^{ij})^2}{2{\sigma_y^{ij}}^2})
\end{align}

Instead of thresholding flow estimation with uncertainty, we use it to adjust the search region for subsequent network interaction, as described in later sections. 

\subsection{Feature Extractor}
As the first part of our network, a convolutional neural network (CNN) is used to extract $1/8$ down-sampled initial features $F_A,F_B  \in R^{\frac{H}{8}\times \frac{W}{8}} $ for each image. As is shown in previous works~\cite{luo2020aslfeat,revaud2019r2d2,detone2018superpoint,contextdesc,geodesc,l2net,hardnet,caps}, CNN exhibits strong capability to capture local context and generates high-level features, which can be directly used to perform nearest neighbour matching. However, since these features are processed independently for each image and critical cross view contexts are missed. To enrich features with long range and cross view contexts, the initial features are further fed into our proposed Transformer module for updating.

\subsection{Initialization} Our Transformer-module starts with a fast initialization block, which conducts (1) positional encoding and (2) two-view contexts initialization. 

\smallskip\noindent\textbf{Positional encoding.} As validated in Transformer networks~\cite{sun2021loftr,sarlin2020superglue,jiang2021cotr}, positional encoding is critical in maintaining spatial information for the flattened tokens. Following the same formulation in LoFTR~\cite{sun2021loftr}, 2D sinusoidal signals in different frequencies are used to encode position information and are added to initial features. Specially, we apply normalization when testing resolution differs from training resolution. We provide more details in Appendix A.5. 

\smallskip\noindent\textbf{Two-view contexts initialization.}
At each local attention phase, our network requires regressing an auxiliary flow map as guidance, which requires cross view contexts. To this end, we pass positional embedded features to a light-weight cross attention block. More specifically, These features are downsampled to low resolution $H_0,W_0$ and two global cross attention blocks are used for feature processing. After initialization, the features are upsampled back to original input resolution, denoted as $F_A^1,F_B^1$, and sent to iterative global-local attention blocks for further processing.

\subsection{Global-Local Attention Block}
The basic structure of our Transformer network is global-local attention (\textbf{GLA}) block. As is shown in Figure~\ref{fig:net_arch}, for each GLA block, attention is performed upon a 3-level coarse-to-fine feature pyramid built by strided average pooling. 

For the $i$-th GLA block, \textbf{global attention} is conducted on coarsest downsampled features in resolution $[H_0,W_0]$, while \textbf{local attention} with adaptive span is used to pass message between medium-resolution features in resolution $[\frac{H}{16},\frac{W}{16}]$ and fine level features in resolution $[\frac{H}{8},\frac{W}{8}]$. Note that we keep the coarsest resolution as a constant, making the complexity of global full attention unaffected by input size. Retrieved messages $M^c,M^m,M^f$ from coarse/medium/fine level are upsampled to same $[\frac{H}{8},\frac{W}{8}]$ resolution, concatenated and fused with an MLP to update source features.
\begin{align}
   M=&\textbf{MLP}(M^c||M^m||M^f), \\
   F^{i+1}&=\textbf{FFN}(M,F^i).
\end{align}
The FFN in our network is defined as
\begin{align}
\textbf{FFN}(M,F)=F+\textbf{LN}(\textbf{Conv}_3(F||M)).
\end{align}
\textbf{LN} stands for layer normalization. Specially, we adopt a $3\times3$ convolution $\textbf{Conv}_3$ in FFN for locality modeling, which compensate for the absence of self attention within each feature map. Empirically, we find $3\times3$ convolution in FFN works better than the combination of linear projection FFN and self attention, more details can be found in Appendix A.5.
\newline

\begin{figure}[t]
	\centering
	\includegraphics[width=0.98\textwidth]{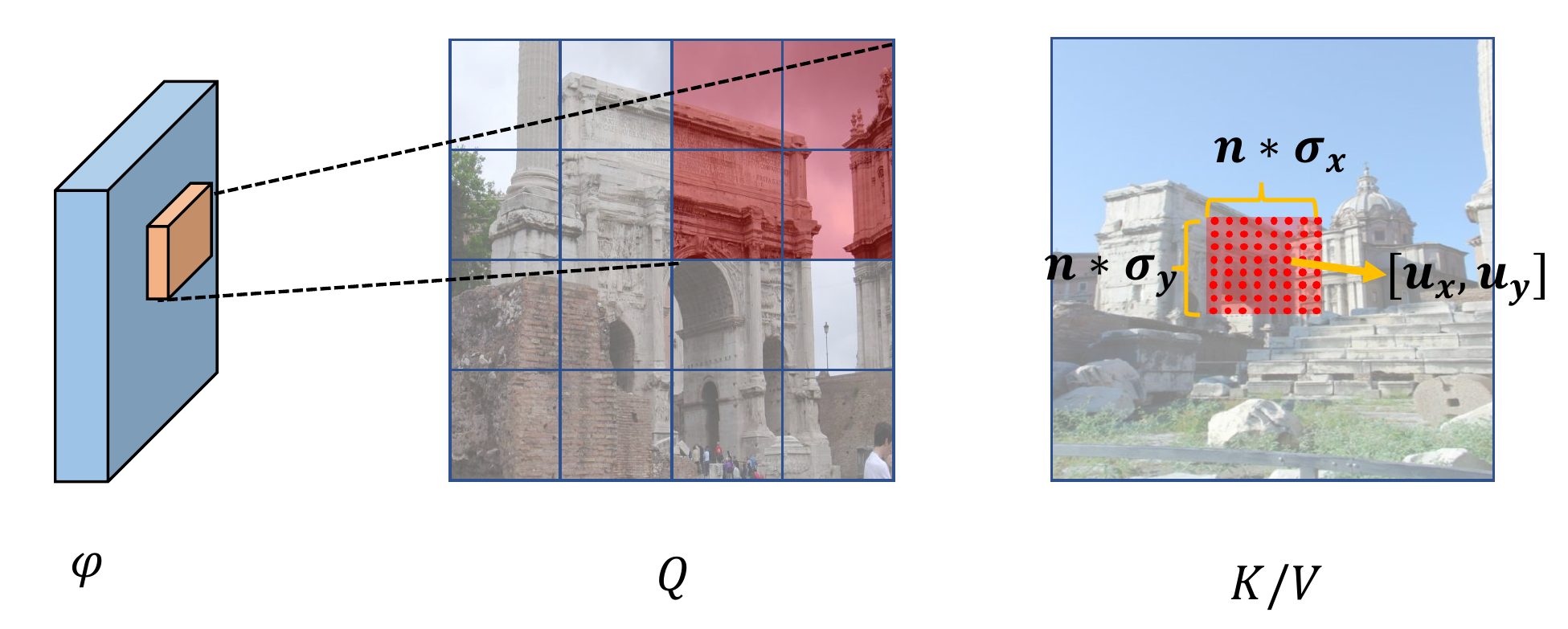}
	\caption{Illustration Local cross attention. Query map $Q$ are partitioned into cells in size $S \times S$($S=2$ in this case), retrieving prediction from flow map $\phi$ and generate attention span. Here we only show attention span for one cell(marked in red).}
	\label{fig:local attention}
\end{figure} 

\smallskip\noindent\textbf{Local cross attention with adaptive attention span.} To facilitate fine-grained attention with modest cost, we adopt local attention on medium and fine level feature maps, where attention span focuses on the neighbourhood regions around current correspondences estimation. 

A key problem for local attention is how to define the size of neighbourhood region. A naive approach is to define neighbourhood with a fixed radius $r$ for all pixels, neglecting the fact that the optimal attention span for different regions varies. For instance, it is sufficient to match regions with distinctive features using small contexts, while regions that are harder to match require larger contexts. Instead of using fixed attention span for all pixels, we propose to adaptively adjust the attention span according to the uncertainty of flow estimation. This design lets each area balance their local receptive fields with uncertainty awareness. Regions with high confidence in flow estimation can sharply focus on a small region for fine level matching, while larger contexts are extracted in low confidence areas for better convergence.

Formally, for the $i$-th GLA block, we first regress flow maps $\phi_A^i,\phi_B^i$ from input features $F_A^i,F_B^i$ in fine level with an MLP, while the medium level flow map are obtained by strided average pooling. For each scale level, we partition the corresponding query map $Q$ into cells with size $S \times S$. For each cell, we use the mean flow estimation to generate a rectangle region upon $K,V$ map and uniformly sample a fixed number of tokens. Attention is performed between each cell and the sampled tokens. The detailed process is defined in Algorithm~\ref{local_algorithm}. An illustration for local attention is given in Fig.~\ref{fig:local attention}.  Since number of sampled tokens for each location is fixed, the whole process retains linear complexity.

\begin{algorithm}[t]
\caption{Local Cross Attention}
\label{local_algorithm} 
\hspace*{0.02in} {\bf Input:}
$Q,K,V  \in R^{H \times W \times C},\phi \in R^{H \times 4}$, span coefficient $n$, sample number $w$, window size $S$ \\
\hspace*{0.02in} {\bf Output:}
Retrieved message $M \in R^{H\times W \times C}$
\begin{algorithmic}[1]
\State Partition $Q$ into cells set $Q_p$ with window size $S \times S$, there will be $\frac{H}{S} \times \frac{W}{S}$ cells in total
\State $M=[\text{~}]$
\For{each cell $Q_{pi} \in R^{S^2 \times C}$ in $Q_p$} \textbf{in parallel}
\State Retrieving flow $\phi_p \in R^{S^2 \times 4}$ from flow map $\phi$ according to the location of $Q_{pi}$
\State Let $[u_x,u_y,\sigma_x,\sigma_y]=\overline{\phi_p}=\sum_{j}{\phi_p[j,:]}$
\State Let $\Gamma$ be a rectangle area with center $[u_x,u_y]$, width $n*\sigma_x$ and height $n*\sigma_y$
\State Uniformly sample $w^2$ tokens in $\Gamma$ region     from $K,V$, denoted as $K_\Gamma,V_\Gamma \in R^{w^2\times C}$
\State Attention $m_i=\text{Att}(Q_{pi},K_\Gamma,V_\Gamma)$
\State Append $m_i$ to $M$
\EndFor
\State Reshaping $M$ into $R^{H\times W \times C}$
\State \Return $M$

\end{algorithmic}
\end{algorithm}

\subsection{Matches Determination}

We inherit the scheme in LoFTR~\cite{sun2021loftr} to generate final correspondences, including a coarse matching stage and a sub-pixel refinement stage. 

After being updated by $N$ GLA blocks, we flatten the output features into $\tilde{F}_A \in R^{n\times c}, \tilde{F}_B \in R^{m \times c}$ and construct correlation matrix $C=\tau \tilde{F}_A \tilde{F}^T_B \in R^{n \times m}$, where $\tau$ is a temperature parameter and $n,m$ are feature numbers of two images. By applying dual-direction softmax in both column/row dimensions, a score matrix is given by $S=\textbf{softmax}_{row}(C) \cdot \textbf{softmax}_{col}(C)$, from which we retain coarse-level matches $M_c$ by mutual nearest neighbour (MNN) and filtering scores below a certain threshold $\theta$. The coarse matches $M_c$ are further fed into a correlation-based refinement block, which is the same with LoFTR~\cite{sun2021loftr}, to obtain the final matching results.

\subsection{Loss Formulation}

We formulate the final loss from three parts, (1) coarse matches loss $L_c$, (2) fine-level loss $L_f$ and (3) flow estimation loss $L_{flow}$
\begin{align}
L= L_c + L_f  +\alpha L_{flow}.
\end{align}
For coarse level loss $L_c$, the ground truth matches $M_{gt}$ is determined by reprojection using depth and camera poses in datasets. We supervise the dual-softmax score matrix $S$ with cross entropy loss
\begin{align}
    L_c=-\frac{1}{|M_{gt}|}\sum_{(i,j)\in M_{gt}} \text{log}(S(i,j)).
\end{align}
The fine-level loss is supervised directly with L2-distance between each refined coordinates $M_f(i,j)$ and ground truth reprojection coordinates, which are further normalized by the coordinate variance. 

For flow estimation supervision, we minimize the log-likelihood for each estimated distribution. Formally, given flow estimation $\Phi$ from each layer and ground truth flow $D^{gt}$, $L_{flow}$ is defined as
\begin{align}
    L_{flow}=-\frac{1}{|D^{gt}|}\sum_{ij} log(P(D^{gt}_{ij}|\Phi_{ij})).
\end{align}
In our implementation, this log-likelihood formulation can be further substituted and decomposed into a more compact form, which is elaborated in Appendix B.

\subsection{Implementation Details}
Our network shares the same ResNet-18~\cite{resnet} CNN feature extractor with LoFTR. After feature extraction and flow initialization, we use 4 GLA blocks for updating. For adaptive attention span, we set $n=5$, meaning that using 5 standard deviation to crop local neighbourhood region for each token. We uniformly sample $8 \times 8$ features in each cropped local region.

We train two different models specified for indoor and outdoor scenes respectively. Both models are optimized using Adam with learning rate $1\times 10^{-3}$ for 30 epochs  on 8 V-100 GPUs. Indoor model is trained on ScanNet~\cite{dai2017scannet} dataset with batch size 24, where the training consumes 5 days. Outdoor model is trained on MegaDepth~\cite{li2018megadepth} with batch size 8, taking 2 days to converge. More details about implementation are introduced in Appendix A.3.

\section{Experiments}

In this section, we demonstrate the performance of our method on two-view pose estimation and visual localization tasks, among both indoor and outdoor scenes. Besides, we conduct ablation study to validate the effectiveness of key design components of our method.

\subsection{Two-view Pose Estimation}
\label{sec:two_view_eval}
We resort to two popular datasets, ScanNet~\cite{dai2017scannet} and MegaDepth~\cite{li2018megadepth}, introduced below, to demonstrate the matching ability of our method in indoor scenes and outdoor scenes, respectively. We also provide additional results on YFCC100M~\cite{yfcc} and Image Matching Challenge(IMC) 2022 in Appendix C.

\smallskip\noindent\textbf{Indoor two-view matching dataset.}
ScanNet dataset~\cite{dai2017scannet} is composed of 1613 sequences, each of which contains RGB images exposing large view changes and repetitive or textureless patterns, with ground-truth depth maps and camera poses associated. For fair comparison, we follow the same training and testing protocols used by SuperGlue~\cite{sarlin2020superglue} and LoFTR~\cite{sun2021loftr}, where 230M and 1.5K image pairs are sampled for training and testing, respectively. In congruent with LoFTR, we resize all test images to $480\times640$.

\smallskip\noindent\textbf{Outdoor two-view matching dataset.}
MegaDepth~\cite{li2018megadepth} consists of 196 3D reconstructions from 1M Internet images, whose camera poses and depth maps are initially computed from COLMAP~\cite{schonberger2016structure} and then refined as ground-truth. We perform two view pose estimation on 1.5k testing pairs. All test images are resized so that their longest dimension is 1152.

\smallskip\noindent\textbf{Evaluation protocols.} Following previous works~\cite{sarlin2020superglue,sun2021loftr}, we train and evaluate our method separately on the two datasets. Two-view pose is recovered by solving essential matrix from correspondences produced, while  pose accuracy is measured by AUC at multiple error thresholds (5$\degree$, 10$\degree$ and 20$\degree$). A pose is only considered accurate if both its angular rotation error and translation error is under a certain threshold compared with ground-truth poses. 

\smallskip\noindent\textbf{Comparative methods.} We compare the proposed method with 1) detector-based approaches, including SuperGlue~\cite{sarlin2020superglue} and SGMNet~\cite{chen2021sgm} that are equipped with SuperPoint(SP)~\cite{detone2018superpoint} as local feature extractor, 2) detector-free approaches, including DRC-Net~\cite{li2020dual}, PDC-Net~\cite{truong2021pdc,pdcnet+}, LoFTR~\cite{sun2021loftr}, QuadTree Attention~\cite{tang2022quadtree}, MatchFormer~\cite{wang2022matchformer} and DKM~\cite{dkm}. 

\begin{figure}[t]
	\centering
	\includegraphics[width=0.98\textwidth]{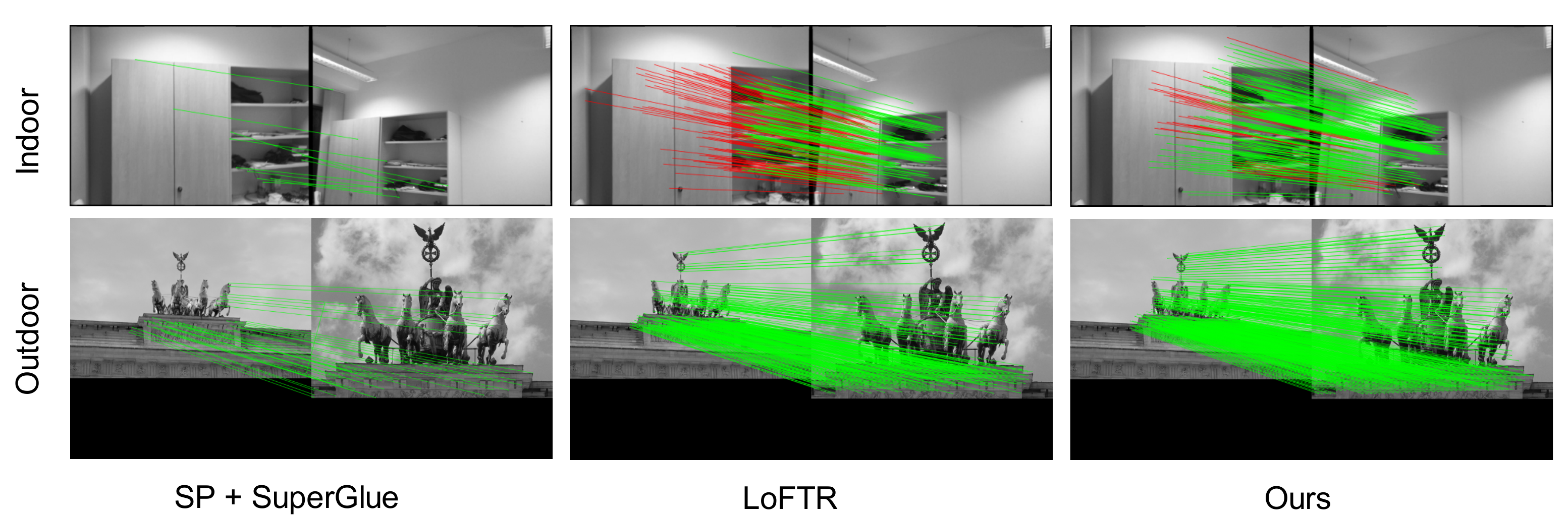}
	\caption{Qualitative results of dense matching in different scenarios. }
	\label{fig:match_vis}
\end{figure}

\smallskip\noindent\textbf{Results.} As presented in Table~\ref{scannet} and Table~\ref{megadepth}, our method consistently achieves the best accuracy in both indoor and outdoor scenes. Visualization in Figure~\ref{fig:match_vis} qualitatively demonstrates our method performance against other matches. More visualizations are provided in Appendix D.

\begin{table}[!ht]
\centering
\begin{minipage}[h]{0.48\linewidth}\centering
\caption{Two-view pose estimation results on ScanNet dataset~\cite{dai2017scannet} in indoor scenes.}
\label{scannet}
\resizebox{0.98\textwidth}{!}{
\begin{tabular}{l>{\centering\arraybackslash}m{1.2cm}>{\centering\arraybackslash}m{1.2cm}>{\centering\arraybackslash}m{1.2cm}}
\Xhline{1pt}
\noalign{\smallskip}
\multirow{2}{*}{\textbf{Method}}  & \multicolumn{3}{c}{\textbf{Pose Estimation AUC}} \\
\noalign{\smallskip}
\cline{2-4}
\noalign{\smallskip}
& @$5\degree$ & @$10\degree$ & @$20\degree$ \\
\noalign{\smallskip}
\Xhline{1pt}
\noalign{\smallskip}
\textit{SP~\cite{detone2018superpoint}+SuperGlue~\cite{sarlin2020superglue}} & 16.2 & 33.8 & 51.8 \\
\textit{SP~\cite{detone2018superpoint}+SGMNet~\cite{chen2021sgm}} & 15.4 & 32.1 & 48.3 \\
\hline
\textit{DRC-Net~\cite{li2020dual}}  & 7.7 & 17.9 & 30.5 \\
\textit{PDC-Net+(H)}~\cite{pdcnet+} & 20.2  & 39.4 & 57.1 \\
\textit{LoFTR~\cite{sun2021loftr}} & 22.0  & 40.8 & 57.6\\
\textit{QuadTree~\cite{tang2022quadtree}} & 24.9  & 44.7 & 61.8\\
\textit{MatchFormer~\cite{wang2022matchformer}} & 24.3  & 43.9 & 61.4 \\
\textit{DKM~\cite{dkm}} & 24.8  & 44.4 & 61.9 \\
\textit{\textbf{Ours}} & \textbf{25.6} & \textbf{46.0} & \textbf{63.3} \\
\Xhline{1pt}
\end{tabular}
}
\end{minipage}\hfill
\begin{minipage}[h]{0.48\linewidth}\centering
\caption{Two-view pose estimation results on MegaDepth dataset~\cite{li2018megadepth} in outdoor scenes.}
\label{megadepth}
\resizebox{0.98\textwidth}{!}{
\begin{tabular}{l>{\centering\arraybackslash}m{1.2cm}>{\centering\arraybackslash}m{1.2cm}>{\centering\arraybackslash}m{1.2cm}}
\Xhline{1pt}
\noalign{\smallskip}
\multirow{2}{*}{\textbf{Method}}  & \multicolumn{3}{c}{\textbf{Pose Estimation AUC}} \\
\noalign{\smallskip}
\cline{2-4}
\noalign{\smallskip}
& @$5\degree$ & @$10\degree$ & @$20\degree$ \\
\noalign{\smallskip}
\Xhline{1pt}
\noalign{\smallskip}
\textit{SP~\cite{detone2018superpoint}+SuperGlue~\cite{sarlin2020superglue}} & 42.2 & 61.2 & 75.9 \\
\textit{SP~\cite{detone2018superpoint}+SGMNet~\cite{chen2021sgm}} & 40.5 & 59.0 & 73.6 \\ 
\hline
\textit{DRC-Net~\cite{li2020dual}}  & 27.0 & 42.9 & 58.3 \\
\textit{PDC-Net+(H)}~\cite{pdcnet+} & 43.1  & 61.9 & 76.1 \\
\textit{LoFTR~\cite{sun2021loftr}} & 52.8  & 69.2 & 81.2\\
\textit{QuadTree~\cite{tang2022quadtree}} & 54.6  & 70.5 & 82.2\\
\textit{MatchFormer~\cite{wang2022matchformer}} & 53.3  & 69.7 & 81.8\\
\textit{DKM~\cite{dkm}} & 54.5  & 70.7 & 82.3\\
\textit{\textbf{Ours}} & \textbf{55.3} & \textbf{71.5} &  \textbf{83.1}\\
\Xhline{1pt}
\end{tabular}
}
\end{minipage}
\end{table}

\subsection{Visual Localization}
Apart from evaluation on two-view pose estimation task, we further integrate our network into a visual localization pipeline, and use two popular datasets, InLoc~\cite{taira2018inloc} and Aachen Day-Night v1.1~\cite{zhang2021reference,sattler2012imageBMCV,Sattler2018CVPR} datasets, to demonstrate performance on multi-view matching in indoor scenes and outdoor scenes, respectively. 

\smallskip\noindent\textbf{Indoor localization dataset.} InLoc dataset~\cite{taira2018inloc} contains a database of $9,972$ RGBD indoor images that are geometrically registered to form the reference scene model, while $329$ RGB query images are provided for visual localization, annotated with manually verified camera poses. Great challenge is posed in matching textureless or repetitive patterns under large perspective differences.

\smallskip\noindent\textbf{Outdoor localization dataset.} Aachen Day-Night v1.1 dataset~\cite{zhang2021reference} depicts a city whose reference scene model is built upon $6,697$ day-time images. For visual localization, the dataset provides another $824$ day-time images and $191$ night-time images as queries. Great challenge is posed in identifying correspondences from, in particular, night-time images under extremely large illumination changes.

\smallskip\noindent\textbf{Evaluation protocols.} We follow the instructions from Long-Term Visual Localization Benchmark~\cite{toft2020long} to compute query poses. For both datasets, we use pre-trained HLoc~\cite{sarlin2019coarse} to retrieve candidate pairs, and recover camera poses with the model trained on MegaDepth dataset following SuperGlue~\cite{sarlin2020superglue} and LoFTR~\cite{sun2021loftr}. More details on localization pipeline are elaborated in Appendix A.4.

\smallskip\noindent\textbf{Results.} On InLoc dataset, as shown in Table~\ref{inloc}, our methods achieves overall best results compared with multiple comparative methods. On Aachen V1.1, as shown in Table~\ref{aachen}, our method outperforms all other methods except SuperGlue. We partially ascribe this to the fact that we use only coarse matches for database reconstruction (see Appendix A.4.), causing localization error that harms pose estimation. In general, our method generalizes well in practical pipelines.
\begin{table}[!ht]
\centering
\begin{minipage}[h]{0.48\linewidth}\centering
\caption{Visual localization results on InLoc dataset~\cite{taira2018inloc}.}
\label{inloc}
\resizebox{0.98\textwidth}{!}{ 
\begin{tabular}{l>{\centering\arraybackslash}m{3cm}>{\centering\arraybackslash}m{3cm}}
\Xhline{1pt}
\noalign{\smallskip}
\multirow{2}{*}{\textbf{Method}}  & \textbf{DUC1} & \textbf{DUC2} \\
\noalign{\smallskip}
\cline{2-3}
\noalign{\smallskip}
 & \multicolumn{2}{c}{(0.25m,2$\degree$) / (0.5m,5$\degree$) / (1m,10$\degree$)} \\
\noalign{\smallskip}
\Xhline{1pt}
\noalign{\smallskip}
\textit{HLoc~\cite{sarlin2019coarse} + SP~\cite{detone2018superpoint}+SuperGlue~\cite{sarlin2020superglue}}  & 49.0 / 68.7 / 80.8 & 53.4 / \textbf{77.1} / 82.4 \\
\textit{HLoc~\cite{sarlin2019coarse}  + LoFTR~\cite{sun2021loftr}} & 47.5 / 72.2 / 84.8 & 54.2 / 74.8 / \textbf{85.5} \\
\textit{HLoc~\cite{sarlin2019coarse}  + Ours} & \textbf{51.5} / \textbf{73.7} / \textbf{86.4} & \textbf{55.0} / 74.0 / 81.7  \\
\Xhline{1pt}
\end{tabular}
}
\end{minipage}\hfill
\begin{minipage}[h]{0.48\linewidth}\centering
\caption{Visual localization results on Aachen V1.1 dataset~\cite{zhang2021reference}.}
\label{aachen}
\resizebox{0.98\textwidth}{!}{
\begin{tabular}{l>{\centering\arraybackslash}m{3cm}>{\centering\arraybackslash}m{3cm}}
\Xhline{1pt}
\noalign{\smallskip}
\multirow{2}{*}{\textbf{Method}}  & \textbf{Day} & \textbf{Night} \\
\noalign{\smallskip}
\cline{2-3}
\noalign{\smallskip}
 & \multicolumn{2}{c}{(0.25m,2$\degree$) / (0.5m,5$\degree$) / (1m,10$\degree$)} \\
\noalign{\smallskip}
\Xhline{1pt}
\multicolumn{3}{l}{\textbf{Localization with matching pairs provided in dataset}} \\
\hline
\textit{R2D2~\cite{revaud2019r2d2}} + NN & - &  71.2 / 86.9 / 98.9 \\
\textit{ASLFeat~\cite{luo2020aslfeat} + NN } & - & 72.3 / 86.4 / 97.9 \\
\textit{SP~\cite{detone2018superpoint}+SuperGlue~\cite{sarlin2020superglue}} & - & 73.3 / 88.0 / 98.4 \\
\textit{SP~\cite{detone2018superpoint}+SGMNet~\cite{chen2021sgm}} & - & 72.3 / 85.3 / 97.9 \\
\hline
\multicolumn{3}{l}{\textbf{Localization with matching pairs generated by HLoc}} \\
\hline
\textit{SP~\cite{detone2018superpoint}+SuperGlue~\cite{sarlin2020superglue}}  & \textbf{89.8} / \textbf{96.1} / \textbf{99.4} & 77.0 / 90.6 / \textbf{100.0} \\
\textit{LoFTR~\cite{sun2021loftr}} & 88.7 / 95.6 / 99.0 & \textbf{78.5}  / 90.6 / 99.0\\
\textit{Ours} & 89.4 / 95.6 / 99.0 & 77.5 / \textbf{91.6} / 99.5  \\
\Xhline{1pt}
\end{tabular}
}
\end{minipage}
\end{table}

\subsection{Ablation Study}
To validate the effectiveness of different design components of our method, we conduct ablation experiments on ScanNet dataset~\cite{dai2017scannet} following the same setting in Section~\ref{sec:two_view_eval}. Specifically, we compare three designs of attention structure:

\begin{itemize}
  \item \textit{Single-Level Attn.}: A design with only global attention at coarsest feature maps without the need of flow estimation. In this design, global context is well captured, whereas essential locality in motion smoothness is omitted and fine-grained message exchange becomes difficult.
  \item \textit{Multi-Level Attn.}: A design with the hierarchical attention framework proposed in this paper, except that the size of local attention span is fixed to 13 px, i.e., the statistical mean of the adaptive attention span used in our network. 
  \item \textit{Adaptive Span Attn.}: Our full design that enables hierarchical attention with adaptive attention span. By this means, the need of context for different pixels is dynamically decided regarding different matchability.
\end{itemize}

As presented in Table~\ref{ablation}, both hierarchical global-local attention and adaptive attention span improve overall performance by a considerable margin, validating the essentiality of our network designs.

\begin{table}[!ht]
\centering
\begin{minipage}[h]{0.48\linewidth}\centering
\caption{Ablation study on ScanNet dataset~\cite{dai2017scannet}.}
\label{ablation}
\resizebox{0.90\textwidth}{!}{
\begin{tabular}{l>{\centering\arraybackslash}m{1.2cm}>{\centering\arraybackslash}m{1.2cm}>{\centering\arraybackslash}m{1.2cm}}
\Xhline{1pt}
\noalign{\smallskip}
\multirow{2}{*}{\textbf{Method}}  & \multicolumn{3}{c}{\textbf{Pose Estimation AUC}} \\
\noalign{\smallskip}
\cline{2-4}
\noalign{\smallskip}
& @$5\degree$ & @$10\degree$ & @$20\degree$ \\
\noalign{\smallskip}
\Xhline{1pt}
\noalign{\smallskip}
\textit{Single-Level Attn.} & 22.65 & 40.72 & 59.06 \\
\textit{Multi-Level Attn.} & 24.85 &  44.86 & 62.71 \\
\textit{\textbf{Adaptive Span Attn.}} & \textbf{25.61} & \textbf{46.04} & \textbf{63.33} \\
\Xhline{1pt}
\end{tabular}
}
\end{minipage}\hfill%
\begin{minipage}[h]{0.48\linewidth}\centering
\caption{Flow estimation accuracy.}
\label{flow}
\resizebox{0.95\textwidth}{!}{
\begin{tabular}{lcccc}
\Xhline{1pt}
\noalign{\smallskip}
\multirow{2}{*}{\textbf{Stage}}  &  \multirow{2}{*}{\textbf{$<$6px (\%)}} & \multicolumn{3}{c}{\textbf{5$\sigma$ \textbf{(px)}}}  \\
\noalign{\smallskip}
\cline{3-5}
\noalign{\smallskip}
& &  Matchable & Unmatchable & Total \\

\noalign{\smallskip}
\Xhline{1pt}
\noalign{\smallskip}
\textit{Iter\#1} & 69.1 & 9.2 & 19.4& 13.4\\ 
\textit{Iter\#2} & 71.2 & 8.2 & 20.2& 12.5\\
\textit{Iter\#3} & 72.0 & 7.8 & 23.8& 12.6 \\
\textit{Iter\#4} & 72.3 & 7.7 & 27.1& 13.3\\
\Xhline{1pt}
\end{tabular}
}
\end{minipage}
\end{table}
\subsection{Understanding ASpanFormer}

\smallskip\noindent\textbf{Flow estimation.} We analyze the flow estimation through multiple iterations. As shown in Table~\ref{flow}, precision of flow regression is gradually improved as attention iterations are performed and converges after four iterations. 

As for uncertainty estimation, we split all pixels into two categories, matchable and unmatchable pixels, identified by ground-truth camera poses and depths, and report their mean standard deviation ($\sigma$). On one hand, mean $\sigma$ decreases with iterations for matchable pixels, as the network becomes more certain about its flow prediction in later stages. On the other hand, the network gradually increases uncertainty values of unmatchable pixels to prevent over-confidence to a certain region. 

\begin{figure}[t]
	\centering
	\includegraphics[width=0.98\textwidth]{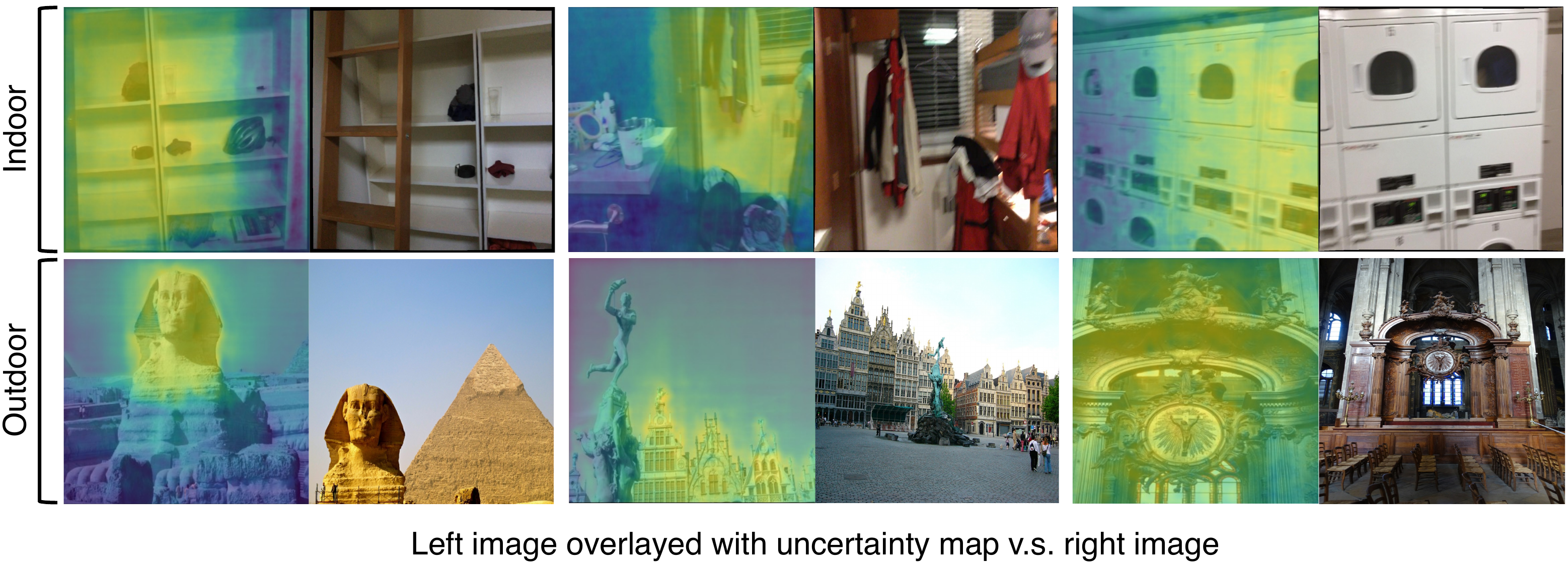}
	\caption{Visualization of uncertainty map which is predicted along with flows, warmer color indicates smaller uncertainties.}
	\label{fig:vis_uncertainty}
\end{figure}

\smallskip\noindent\textbf{Uncertainty map.} In Figure~\ref{fig:vis_uncertainty}, we provide visualization of uncertainty map of flow prediction. Overlapping and non-overlapping regions are firstly distinguished, while uncertainty values in textureless regions are usually larger, indicating context of larger size is required during cross attention.   

\smallskip\noindent\textbf{Runtime evaluation.} We evaluate the runtime of proposed method and compare it with LoFTR~\cite{sun2021loftr} where both methods apply Transformer backend. The runtime speed is tested on 100 randomly sampled ScanNet image pairs ($640\times 480$) with a NVIDIA V100 GPU. Runtime differs only on \textit{Attention Module} compared with LoFTR, as we adopt the same Local Feature \textit{CNN} backbone and coarse-to-fine matching module. As shown in Table~\ref{efficiency}, the proposed method is overall slightly slower than LoFTR due to the more complicated attention operation.
\begin{table}[!ht]
\centering
\caption{Runtime speed evaluated on 640$\times$480 images. }
\label{efficiency}
\resizebox{0.6\textwidth}{!}{ 
\begin{tabular}{l>{\centering\arraybackslash}m{3cm}>{\centering\arraybackslash}m{3cm}}
\Xhline{1pt}
\noalign{\smallskip}
\multirow{2}{*}{\textbf{Stage}}  & \multicolumn{2}{c}{\textbf{Runtime (ms)}} \\
\noalign{\smallskip}
\cline{2-3}
\noalign{\smallskip}
  & LoFTR & Ours \\

\noalign{\smallskip}
\Xhline{1pt}
\noalign{\smallskip}
\textit{Local Feature CNN} & 32.2 &  32.2 \\
\textit{Attention Module} & 24.6 &  40.5\\
\textit{Matching} & 40.9 & 40.8 \\
\textit{Total}  & 97.7 & 113.5\\
\Xhline{1pt}
\end{tabular}
}
\end{table}
\vspace{-1em}

\section{Conclusion}
In this paper, we have proposed a novel Transformer framework based on feature hierarchy, whose attention span is adaptively decided so as to acquire capabilities
to capture both long-term dependencies as well as fine-grained details in local regions. State-of-the-art results validates the effectiveness of our method. With more engineering optimizations, we are looking forward to wider application of our method in real use. 

\appendix
\section*{\centering{Appendix}} 
\renewcommand\thesection{\Alph{section}}

\section{Implementation Details}
In this section, we provide more details about our network implementation. 

\subsection{Network Settings}

We use the same ResNet-18 for initial feature extractor as that in LoFTR, which outputs feature maps in two resolution, $\frac{1}{8}$ and $\frac{1}{2}$. The $\frac{1}{8}$ feature map is passed into our transformer-based network for updating, while the $\frac{1}{2}$ is used in fine matches coordinates refinement. For dual-softmax in coarse matching, we adopt a learnable temperature which is initialized as 10.

We use four GLA blocks to update features. For hierarchical attention, we fix the coarsest feature maps in resolution $H_0,W_0$, where $(H_0,W_0)=(15,20)$ for indoor settings and $(H_0,W_0)=(36,36)$ for outdoor settings. 

\subsection{Flow Regression}

As stated in Sec. 3.4, we use an MLP to regress auxilary flow map in each GLA block. Given D-dimensional feature in pixel, we use MLP with shape (D,64,4) to regress a 4-dimensional feature $f$. For corresponding coordinates $u_x,u_y$, We normalize the first two values with sigmoid function and recover them to the range of image resolution. For the standard variance $\sigma_x,\sigma_y$, we regress the last two values as their logarithm. Formally,

\begin{align}
    [u_x,u_y]=Sigmoid(f[:2])*[H,W],~[\sigma_x,\sigma_y]=exp(f[2:])
\end{align}

where $H,W$ are image height and width.

\subsection{Training Details}

\begin{figure}[t]
	\centering
	\includegraphics[width=0.48\textwidth]{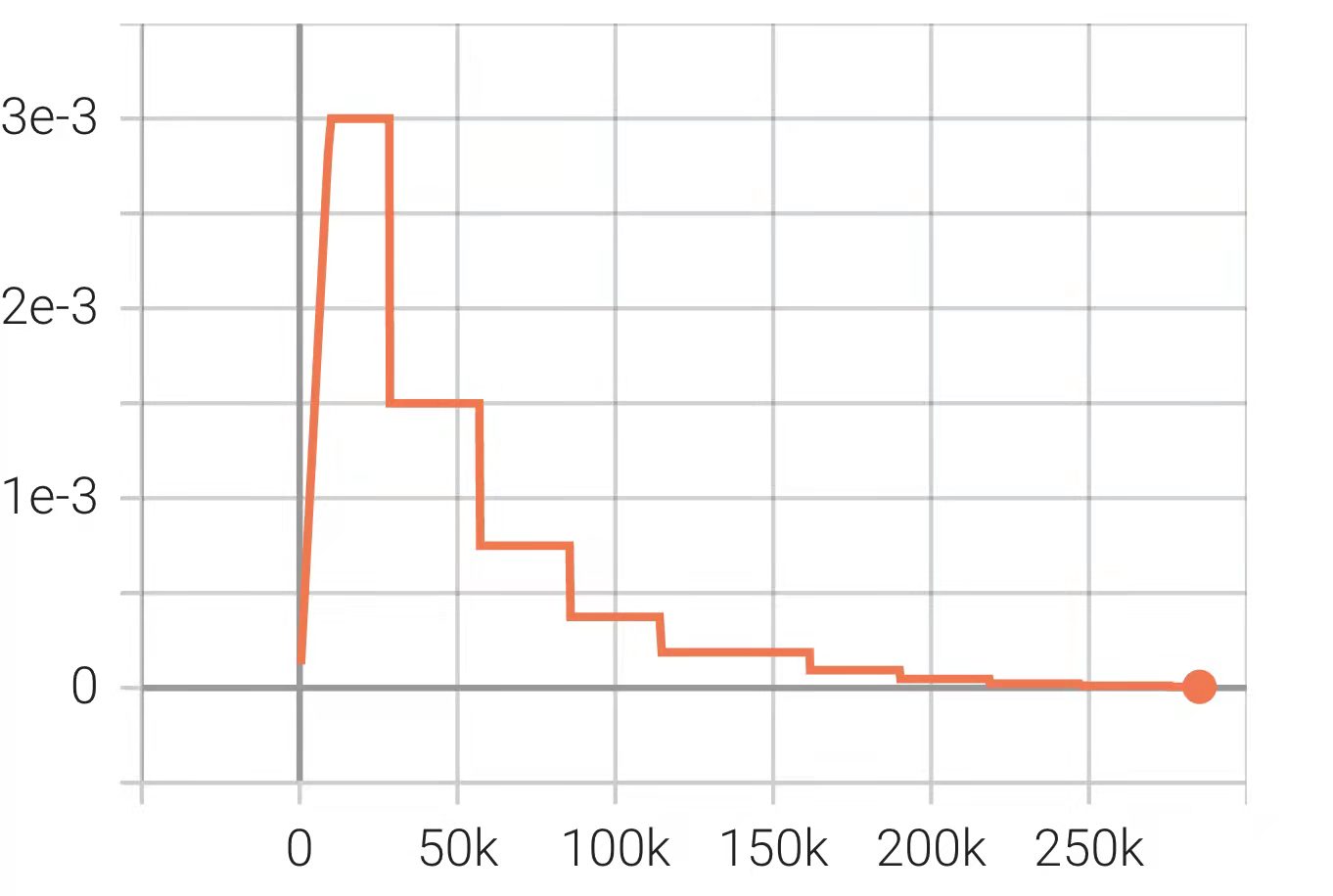}
	\caption{Learning rate curve across iterations. }
	\label{fig:lr}
\end{figure}

For both indoor and outdoor training, we adopt the same muti-step training strategy as that in officially released LoFTR code. More specifically, the learning rate is linearly warmed-up in this first epoch and then halved every two or three epochs. The learning rate curve is illustrated in Fig.~\ref{fig:lr}.

\subsection{Visual Localization Details}

We refer to hierachical localization pipeline (https://github.com/cvg/Hierarchical-Localization) to perform viusal localization experiments on Aachen Day-Night and InLoc datasets.

For Aachen Day-Night, we first triangulate reference models by using only coarse matches across images. We then generate fine level matches between query images and database images, where the database images are taken as left images, so that the fine level matches can be registered to triangulated 3D tracks.

For InLoc dataset, we directly generate fine level matches between query and database images, where the 2D match points on reference images are projected to 3D space through the provided depth map. We omit image pairs with fewer than 25 matches. 

\subsection{Some Effective Designs}
We provide ablations for some additional useful designs in our network: (1) learnable temperature for softmax at each level. (2)Convolution-based FFN. (3) Normalized positional encoding when testing resolution differs from training resolution. An ablation study for these techniques is provided in Tab.~\ref{tricks} and Tab.~\ref{pe}.

\smallskip\noindent\textbf{Learnable Temperature.} As stated in Sec.~3.4, message $M^f,M^m,M^c$ are computed from different levels of feature maps through global or local attention, where softmax are applied to tokens in different numbers. A concern about softmax is the that the number of tokens largely affect the final distribution. To balance the impact of different token number in global/local attention, we adopt three learnable temperature parameters $\tau_f,\tau_m,\tau_c$ for softmax in fine, medium and coarse level features respectively.   

\smallskip\noindent\textbf{Convolutional FFN.} As shown in Sec. 3, our networks is fully based on cross attention for cross-view message passing, while self attention is absent. Deviating from common practice that employs self attention for intra-image message passing, we find in our experiment that adopting $3 \times 3$ convolution in FFN to replace self attention and MLP-based FFN leads to better overall performance. 

\smallskip\noindent\textbf{Normalized Positional Encoding.} Positional encoding (PE) in LoFTR is defined as,

\begin{center}
$ PE^i(x,y)=\left\{
\begin{aligned}
sin(w_k\cdot x),~&i=4k \\
cos(w_k\cdot x),~&i=4k+1 \\
sin(w_k\cdot x),~&i=4k+2 \\
cos(w_k\cdot x),~&i=4k+3
\end{aligned}
\right. $
\end{center}

A concern about this PE is that unseen coordinate will be used in encoding when testing resolution differs from training resolution, which harms the network's capability of precise localization and boundary awareness. To mitigate the issue, we adopt a simple normalization technique,

\begin{align}
     PE_n^i(x,y)&=PE^i(x*\alpha,y*\beta)\\
     \alpha=W_{tain}/&W_{test}, ~\beta=H_{train}/H_{test}
\end{align}

where $W/H_{train/test}$ are width/height of training/testing image. We find this normalization boost the performance of our method when training/testing image resolution differ. Aligning testing/training PE is especially critical for precise flow prediction, since it relies on PE to regress flow coordinate.

In Tab.~\ref{pe}, we provide ablation study results for normalized positional encoding (NPE). The results are obtained on MegaDepth dataset with all images resized to 1152 resolution, while the models are trained in 832 resolution.

\begin{table}[!ht]
\centering
\begin{minipage}[h]{0.48\linewidth}\centering
\caption{Ablations on network designs on ScanNet~\cite{dai2017scannet} dataset. SA+MLP-FFN, means adopting 1/4 downsampled self attention after each GLA block and replacing all $3 \times 3$ conv in FFN of both self/cross attention with MLP.}
\label{tricks}
\resizebox{0.98\textwidth}{!}{
\begin{tabular}{l>{\centering\arraybackslash}m{1.2cm}>{\centering\arraybackslash}m{1.2cm}>{\centering\arraybackslash}m{1.2cm}}
\Xhline{1pt}
\noalign{\smallskip}
\multirow{2}{*}{\textbf{Method}}  & \multicolumn{3}{c}{\textbf{Pose Estimation AUC}} \\
\noalign{\smallskip}
\cline{2-4}
\noalign{\smallskip}
& @$5\degree$ & @$10\degree$ & @$20\degree$ \\
\noalign{\smallskip}
\Xhline{1pt}
\noalign{\smallskip}
\textit{AspanFormer w/o learnable temperature} & 25.0 & 45.7 & 62.3 \\
\textit{AspanFormer w SA+MLP-FFN} & 24.8 & 45.5 & 62.0 \\
\textit{\textbf{AspanFormer}} & \textbf{25.6} & \textbf{46.0} & \textbf{63.3} \\
\Xhline{1pt}
\end{tabular}
}
\end{minipage}\hfill%
\begin{minipage}[h]{0.48\linewidth}\centering
\caption{Ablation study of Normalized Positional Encoding (NPE) on MegaDepth dataset~\cite{li2018megadepth}.}
\label{pe}
\resizebox{0.98\textwidth}{!}{
\begin{tabular}{l>{\centering\arraybackslash}m{1.2cm}>{\centering\arraybackslash}m{1.2cm}>{\centering\arraybackslash}m{1.2cm}>{\centering\arraybackslash}m{1.2cm}}
\Xhline{1pt}
\noalign{\smallskip}
\multirow{2}{*}{\textbf{Method}}  & \multicolumn{3}{c}{\textbf{Pose Estimation AUC}} &  \multirow{2}{*}{\textbf{Flow Acc.}}  \\
\noalign{\smallskip}
\cline{2-4}
\noalign{\smallskip}
& @$5\degree$ &  @$10\degree$ & @$20\degree$ & \\

\noalign{\smallskip}
\Xhline{1pt}
\noalign{\smallskip}
\textit{AspanFormer w/o NPE} & 52.8 & 69.6 & 81.1 & 22.6 \\
\textit{AspanFormer} & 55.3 & 71.5 & 83.1 & 72.3\\
\Xhline{1pt}
\end{tabular}
}
\end{minipage}
\end{table}

\section{Flow Loss}
We formulate flow supervision as most-likelihood estimation for Gaussian distribution $P$.

\begin{align}
    L_{flow}=-\frac{1}{|D^{gt}|}\sum_{ij} log(P(D^{gt}_{ij}|\Phi_{ij}))
\end{align}

where $D_{ij}^{gt}=(x_{ij},y_{ij})$ is the ground truth flow and $\phi_{ij}=(u_x^{ij},u_y^{ij},\sigma_x^{ij},\sigma_y^{ij})$ are predicted parameters at location $(i,j)$. Substituting into Gaussian distribution formula, we have

\begin{align}
    L_{flow}&=-\frac{1}{|D^{gt}|}\sum_{ij} log [\frac{1}{2\pi\sigma_x^{ij}\sigma_y^{ij}}\text{exp}(-{\frac{(x_{ij}-u_x^{ij})^2}{2{\sigma_x^{ij}}^2}-\frac{(y_{ij}-u_y^{ij})^2}{2{\sigma_y^{ij}}^2}})]\\
    &=\frac{1}{|D^{gt}|}\sum_{ij} [log 2\pi+log\sigma_x^{ij}+log\sigma_y^{ij}+\frac{(x_{ij}-u_x^{ij})^2}{2{\sigma_x^{ij}}^2}+\frac{(y_{ij}-u_y^{ij})^2}{2{\sigma_y^{ij}}^2}]
\end{align}

In implementation, we let $w_x^{ij}=log\sigma_x^{ij},w_y^{ij}=log\sigma_y^{ij}$ and omit constant terms, then

\begin{align}
L_{flow}&=\frac{1}{|D^{gt}|}\sum_{ij} [w_x^{ij}+w_y^{ij}+\frac{1}{2}e^{-2w_x^{ij}}(x_{ij}-u_x^{ij})^2+\frac{1}{2}e^{-2w_y^{ij}}(y_{ij}-u_y^{ij})^2]
\end{align}

Intuitively, this loss formulation is a weighted sum of L2-distance between estimated flows and ground truth flows. $w_x^{ij}+w_y^{ij}$ is a regularization term encouraging lower uncertainty. The overall effect of flow loss is to minimize uncertainty and flow estimation error simultaneously.

\section{Additional Quantitative Results}

We provide in this part additional experiment results on YFCC100M dataset and Image Matching Challenge 2022 (IMC 2022) kaggle benchmark.

\subsection{Results on YFCC100M}
YFCC100M contains a collection of internet images across various tourism landmarks. We adopt the test set from 4 selected landmark sequences as is done in previous works~\cite{zhang2019oanet,sarlin2020superglue,chen2021sgm}. 1000 image pairs are sampled from each sequence, which yields 4000 pairs test set in total. We use OpenCV ransac for two-view pose estimation, where the RANSAC threshold for \textbf{all methods} is set to $5\times 10^{-4}$ in normalized image coordinate space. Experiment results are given in Tab.~\ref{yfcc}, where our method outperforms all comparative methods. 

\subsection{Results on Image Matching Challenge 2022}
We submit our method to Image Matching Challenge (IMC) 2022 and report the results in Tab.~\ref{imc}. We resize the input image to a fixed resolution [1472,832] and use OpenCV USAC\_MAGSAC to estimate fundamental matrix, where the RANSAC threshold is set to 0.2 pixel. The results show that our method consistently outperforms other strong comparative baselines.

\begin{table}[!ht]
\centering
\begin{minipage}[h]{0.48\linewidth}\centering
\caption{Two-view pose estimation results on YFCC100M~\cite{yfcc} dataset in outdoor scenes.}
\label{yfcc}
\resizebox{0.98\textwidth}{!}{
\begin{tabular}{l>{\centering\arraybackslash}m{1.2cm}>{\centering\arraybackslash}m{1.2cm}>{\centering\arraybackslash}m{1.2cm}}
\Xhline{1pt}
\noalign{\smallskip}
\multirow{2}{*}{\textbf{Method}}  & \multicolumn{3}{c}{\textbf{Pose Estimation AUC}} \\
\noalign{\smallskip}
\cline{2-4}
\noalign{\smallskip}
& @$5\degree$ & @$10\degree$ & @$20\degree$ \\
\noalign{\smallskip}
\Xhline{1pt}
\noalign{\smallskip}
\textit{SP~\cite{detone2018superpoint}+SuperGlue~\cite{sarlin2020superglue}} & 38.1 & 58.8 & 74.7 \\
\textit{RootSIFT+SGMNet~\cite{chen2021sgm}} & 35.5 & 55.2 & 71.9 \\
\hline
\textit{DRC-Net~\cite{li2020dual}}& 29.5 & 50.1 & 66.8   \\
\textit{PDC-Net+(H)}~\cite{pdcnet+} & 39.1  & 60.1 & 76.5 \\
\textit{LoFTR~\cite{sun2021loftr}} &  42.4 & 62.5 & 77.3\\
\textit{\textbf{Ours}} & \textbf{44.5} & \textbf{63.8} & \textbf{78.4} \\
\Xhline{1pt}
\end{tabular}
}
\end{minipage}\hfill%
\begin{minipage}[h]{0.48\linewidth}\centering
\caption{Two-view pose estimation results on IMC 2022 kaggle benchmark. The Results of MatchFormer and QuadTree attention are reported by the 4th solution on Kaggle discussion forum~\cite{4th_imc}.  }
\label{imc}
\resizebox{0.98\textwidth}{!}{
\begin{tabular}{l>{\centering\arraybackslash}m{1.2cm}>{\centering\arraybackslash}m{1.2cm}}
\Xhline{1pt}
\noalign{\smallskip}
\multirow{2}{*}{\textbf{Method}}  & \multicolumn{2}{c}{\textbf{Pose Estimation mAA}} \\
\noalign{\smallskip}
\cline{2-3}
\noalign{\smallskip}
& Private & Public \\
\noalign{\smallskip}
\Xhline{1pt}
\noalign{\smallskip}
\textit{SP~\cite{detone2018superpoint}+SuperGlue~\cite{sarlin2020superglue}} & 0.724 & 0.728  \\
\textit{LoFTR~\cite{sun2021loftr}} & 0.783  & 0.772 \\
\textit{MatchFormer~\cite{wang2022matchformer}} & 0.783  & 0.774 \\
\textit{QuadTree~\cite{tang2022quadtree}} & 0.817  & 0.812 \\
\textit{\textbf{Ours}} & \textbf{0.838} & \textbf{0.833} \\
\Xhline{1pt}
\end{tabular}
}
\end{minipage}
\end{table}

\section{Additional Visualizations}

\begin{figure}[t]
	\centering
	\includegraphics[width=0.98\textwidth]{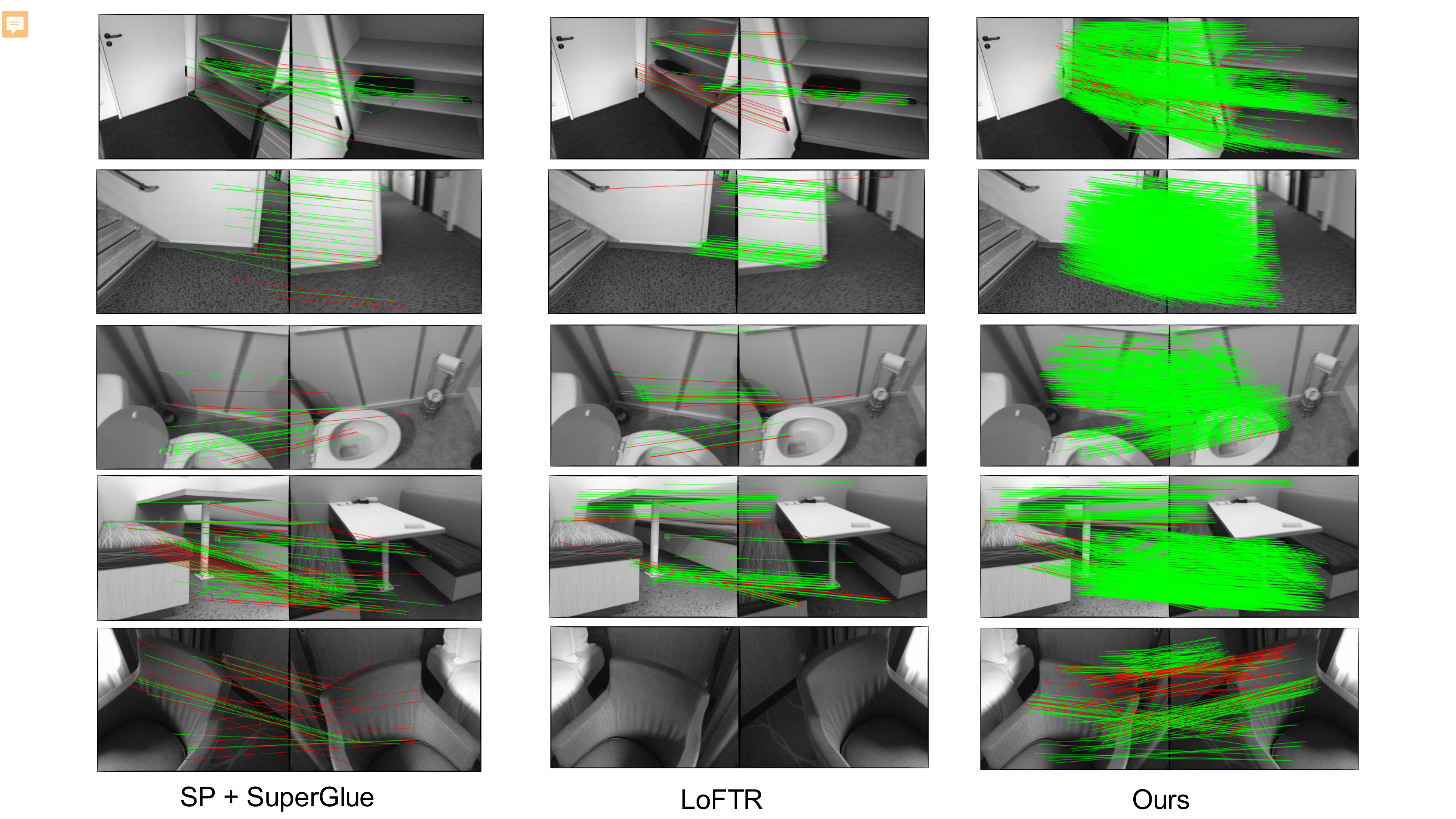}
	\includegraphics[width=0.98\textwidth]{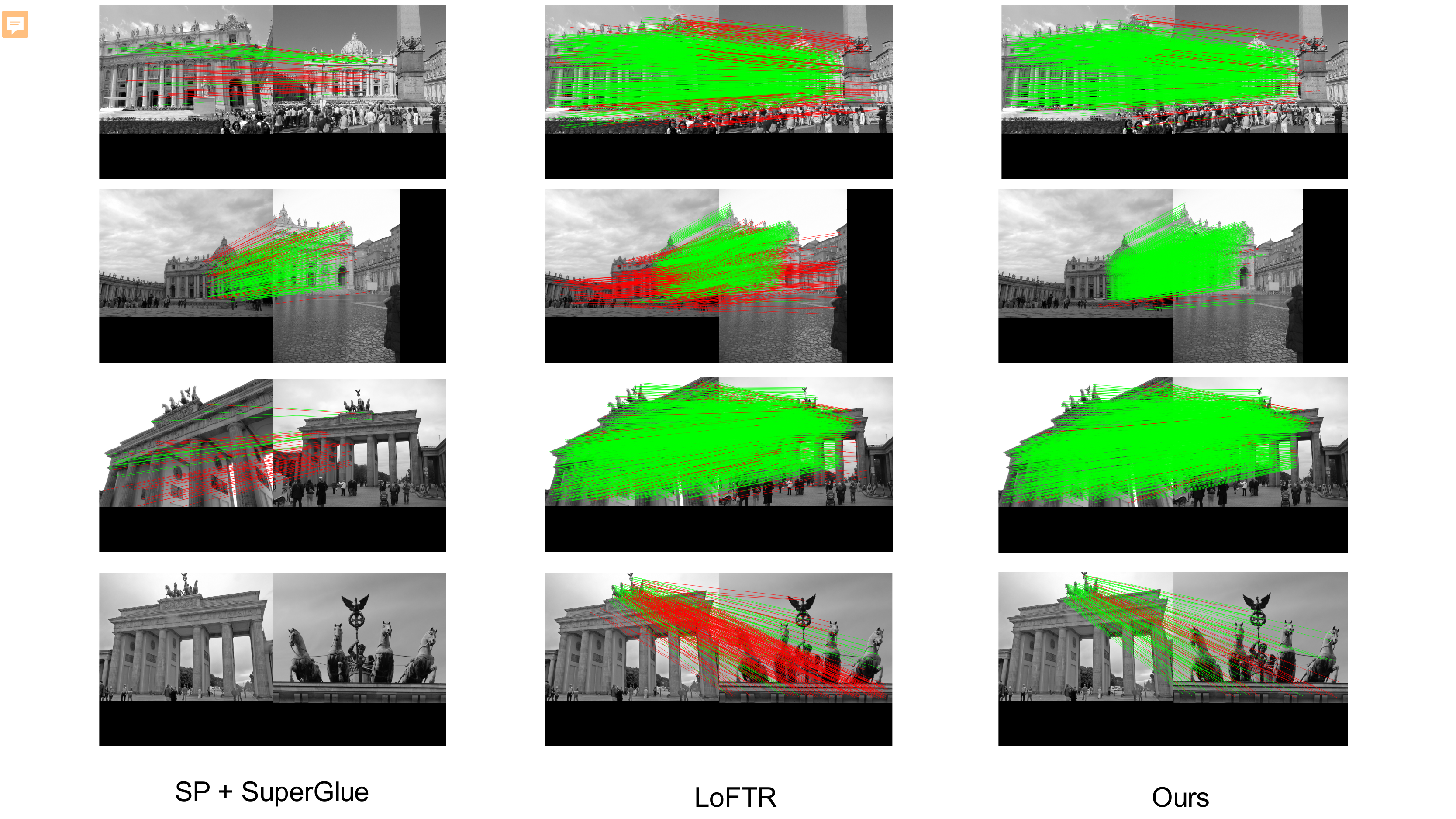}
	\caption{Visualizations of matches obtained through SuperGlue, LoFTR and ASpanFormer(ours). Our methods produces more accurate and denser matches compared with both SOTA sparse and dense matching networks. }
	\label{fig:matches}
\end{figure}

\begin{figure}[t]
	\centering
	\includegraphics[width=0.98\textwidth]{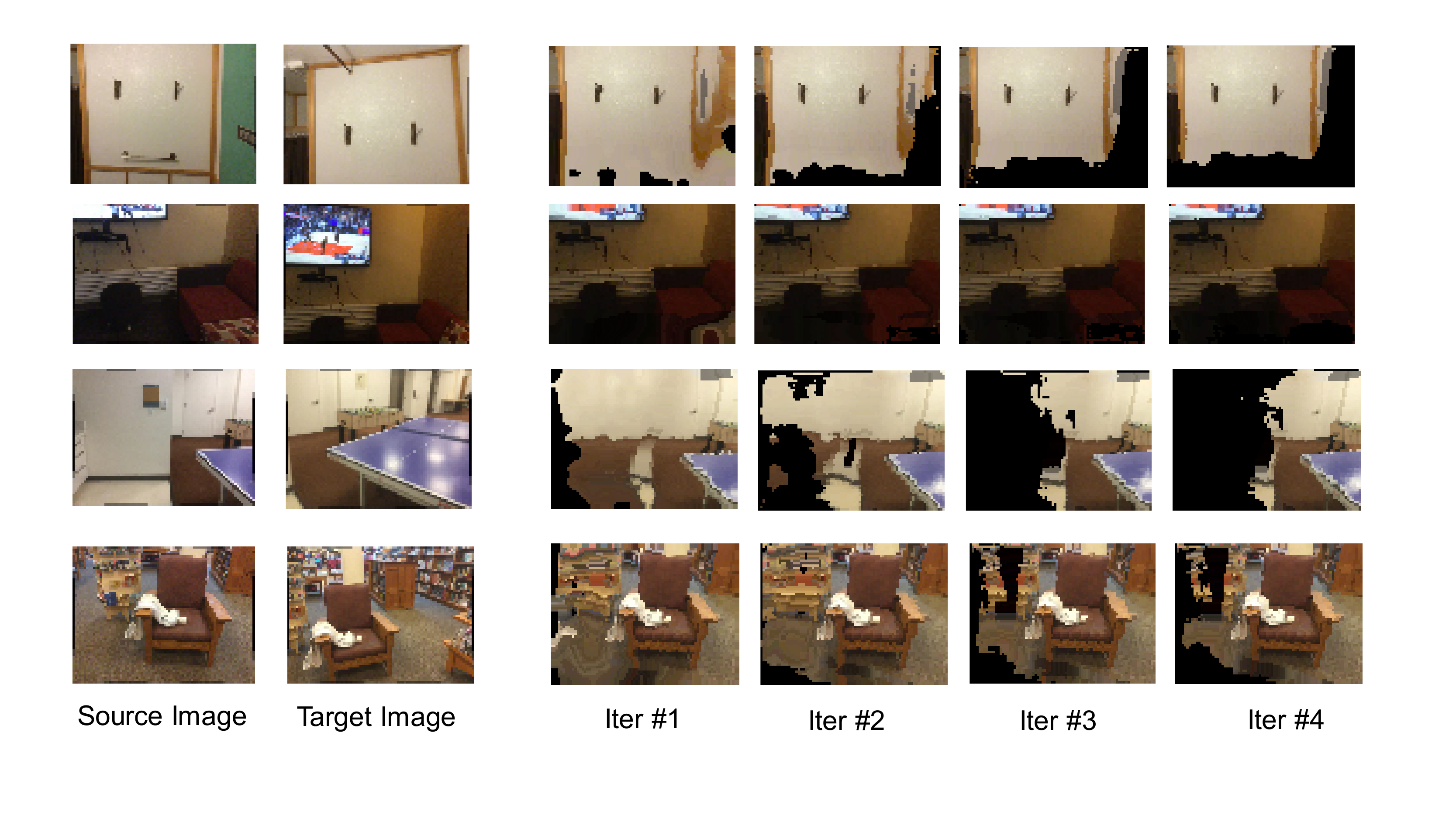}
	\caption{Visualizations of flow prediction across GLA iterations. We filter flow predictions with high uncertainty. Note that the flow map are in $\frac{1}{8}$ ($60 \times 80$) resolution. As more GLA blocks are employed for feature updating, the flow map gradually prune occluded or non-overlap regions.}  
	\label{fig:flow}
\end{figure}

\begin{figure}[t]
	\centering
	\includegraphics[width=0.98\textwidth]{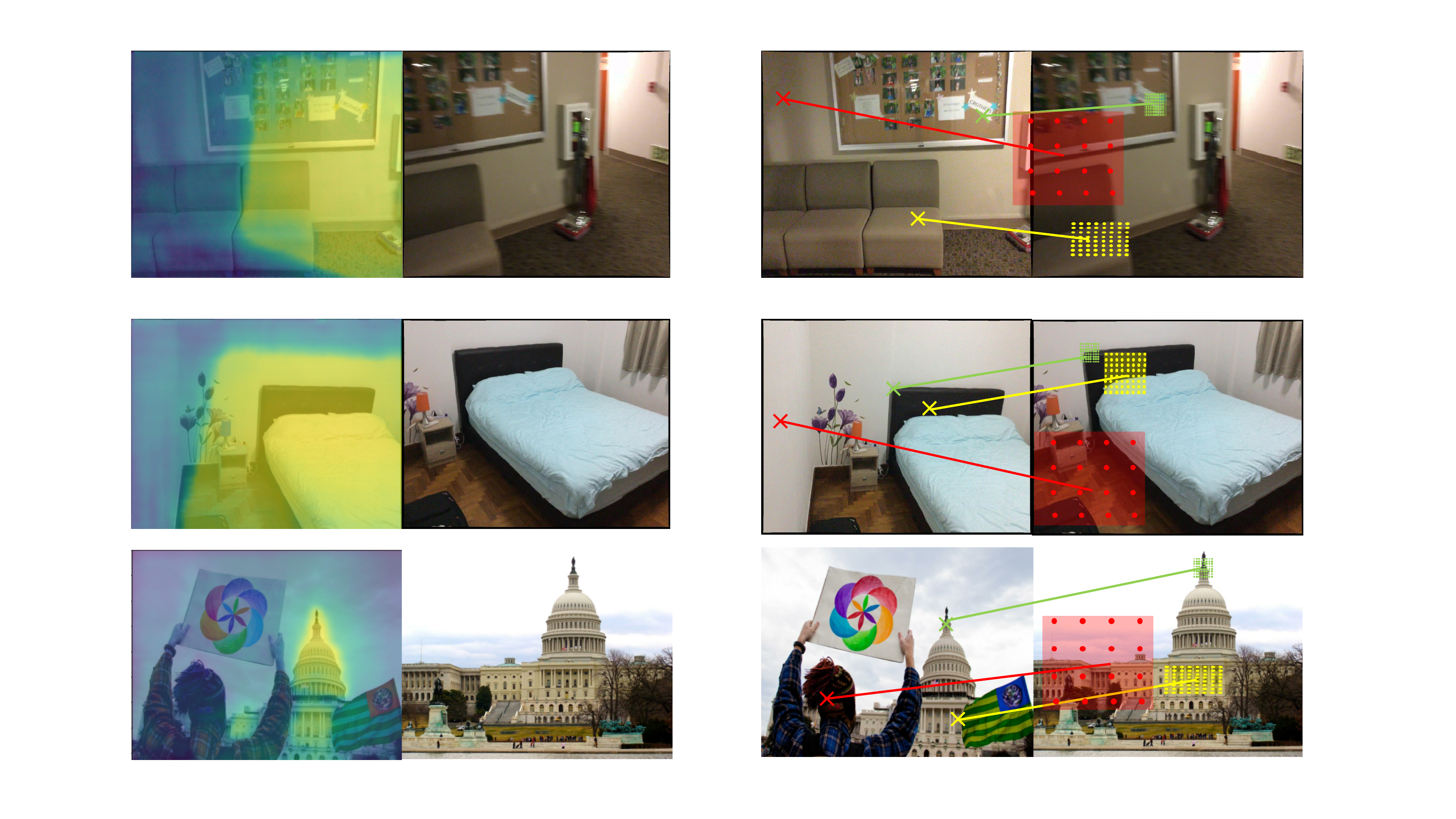}
	\caption{Visualizations of uncertainty heatmap(left) and adaptive attention span(right). Our network sharply focuses on regions with rich and distinctive textures with small attention span, while larger contexts are extracted for the low texture or uncertain regions. Specially, very large attention spans are generated for non-overlapping or occluded areas, preventing falsely focusing on certain regions.}
	\label{fig:span}
\end{figure}

We provide more visualization results in this part. In Fig~\ref{fig:matches}, we provide qualitative comparisons between SuperGlue, LoFTR and our methods. In Fig~\ref{fig:flow}, we provide flow predictions across GLA block iterations. In Fig~\ref{fig:span}, we provide additional visualization of uncertainty heatmap and corresponding adaptive attention spans.

\clearpage

\bibliographystyle{splncs}
\bibliography{egbib}

\begin{thebibliography}{10}

\bibitem{schonberger2016structure}
Schonberger, J.L., Frahm, J.M.:
\newblock Structure-from-motion revisited.
\newblock In: CVPR. (2016)

\bibitem{sfm2}
Resindra, A., Torii, A., Okutomi, M.:
\newblock Structure from motion using dense cnn features with keypoint
  relocalization.
\newblock IPSJ Transactions on Computer Vision and Applications (2018)

\bibitem{sattler2012imageBMCV}
Sattler, T., Weyand, T., Leibe, B., Kobbelt, L.:
\newblock Image retrieval for image-based localization revisited.
\newblock In: BMVC. (2012)

\bibitem{mur2015orb}
Mur-Artal, R., Montiel, J.M.M., Tardos, J.D.:
\newblock {ORB-SLAM}: a versatile and accurate monocular slam system.
\newblock IEEE transactions on robotics (2015)

\bibitem{orbslam2}
Mur-Artal, R., Tardos, J.:
\newblock {ORB-SLAM2}: an open-source slam system for monocular, stereo and
  rgb-d cameras.
\newblock IEEE Transactions on Robotics (2016)

\bibitem{lowe2004distinctive}
Lowe, D.G.:
\newblock Distinctive image features from scale-invariant keypoints.
\newblock IJCV (2004)

\bibitem{rublee2011orb}
Rublee, E., Rabaud, V., Konolige, K., Bradski, G.R.:
\newblock {ORB}: An efficient alternative to sift or surf.
\newblock In: ICCV. (2011)

\bibitem{revaud2019r2d2}
Revaud, J., Weinzaepfel, P., De~Souza, C., Pion, N., Csurka, G., Cabon, Y.,
  Humenberger, M.:
\newblock {R2D2}: repeatable and reliable detector and descriptor.
\newblock In: NeurIPS. (2019)

\bibitem{detone2018superpoint}
DeTone, D., Malisiewicz, T., Rabinovich, A.:
\newblock Superpoint: Self-supervised interest point detection and description.
\newblock In: CVPRW. (2018)

\bibitem{luo2020aslfeat}
Luo, Z., Zhou, L., Bai, X., Chen, H., Zhang, J., Yao, Y., Li, S., Fang, T.,
  Quan, L.:
\newblock Aslfeat: Learning local features of accurate shape and localization.
\newblock In: CVPR. (2020)

\bibitem{d2net}
Dusmanu, M., Rocco, I., Pajdla, T., Pollefeys, M., Sivic, J., Torii, A.,
  Sattler, T.:
\newblock D2-net: A trainable cnn for joint description and detection of local
  features.
\newblock In: CVPR. (2019)

\bibitem{contextdesc}
Luo, Z., Shen, T., Zhou, L., Zhang, J., Yao, Y., Li, S., Fang, T., Quan, L.:
\newblock Contextdesc: Local descriptor augmentation with cross-modality
  context.
\newblock In: CVPR. (2019)

\bibitem{sun2021loftr}
Sun, J., Shen, Z., Wang, Y., Bao, H., Zhou, X.:
\newblock Loftr: Detector-free local feature matching with transformers.
\newblock In: CVPR. (2021)

\bibitem{jiang2021cotr}
Jiang, W., Trulls, E., Hosang, J., Tagliasacchi, A., Yi, K.M.:
\newblock {COTR}: Correspondence transformer for matching across images.
\newblock In: CVPR. (2021)

\bibitem{truong2021pdc}
Truong, P., Danelljan, M., Gool, L.V., Timofte, R.:
\newblock Learning accurate dense correspondences and when to trust them.
\newblock In: CVPR. (2021)

\bibitem{ncnet}
Rocco, I., Cimpoi, M., Arandjelovi, R., Torii, A., Pajdla, T., Sivic, J.:
\newblock Neighbourhood consensus networks.
\newblock In: NeurIPS. (2018)

\bibitem{li2020dual}
Li, X., Han, K., Li, S., Prisacariu, V.:
\newblock Dual-resolution correspondence networks.
\newblock In: NeurIPS. (2020)

\bibitem{truong2020glu}
Truong, P., Danelljan, M., Timofte, R.:
\newblock {GLU-Net}: Global-local universal network for dense flow and
  correspondences.
\newblock In: CVPR. (2020)

\bibitem{houghnet}
Min, J., Cho, M.:
\newblock Convolutional hough matching networks.
\newblock In: CVPR. (2021)

\bibitem{ransasflow}
Shen, X., Darmon, F., Efros, A., Aubry, M.:
\newblock Ransac-flow: Generic two-stage image alignment.
\newblock In: ECCV. (2020)

\bibitem{tang2022quadtree}
Tang, S., Zhang, J., Zhu, S., Tan, P.:
\newblock Quadtree attention for vision transformers.
\newblock In: ICLR. (2021)

\bibitem{rocco2020efficient}
Rocco, I., Arandjelovi{\'c}, R., Sivic, J.:
\newblock Efficient neighbourhood consensus networks via submanifold sparse
  convolutions.
\newblock In: ECCV. (2020)

\bibitem{vaswani2017attention}
Vaswani, A., Shazeer, N., Parmar, N., Uszkoreit, J., Jones, L., Gomez, A.N.,
  Kaiser, {\L}., Polosukhin, I.:
\newblock Attention is all you need.
\newblock In: NeurIPS. (2017)

\bibitem{dosovitskiy2020image}
Dosovitskiy, A., Beyer, L., Kolesnikov, A., Weissenborn, D., Zhai, X.,
  Unterthiner, T., Dehghani, M., Minderer, M., Heigold, G., Gelly, S.,  et~al.:
\newblock An image is worth 16x16 words: Transformers for image recognition at
  scale.
\newblock In: ICLR. (2020)

\bibitem{katharopoulos2020transformers}
Katharopoulos, A., Vyas, A., Pappas, N., Fleuret, F.:
\newblock Transformers are rnns: Fast autoregressive transformers with linear
  attention.
\newblock In: ICML. (2020)

\bibitem{hardnet}
Mishchuk, A., Mishkin, D., Radenović, F., Matas, J.:
\newblock Working hard to know your neighbor's margins:local descriptor
  learning loss.
\newblock In: NeurIPS. (2017)

\bibitem{l2net}
Tian, Y., Fan, B., Wu, F.:
\newblock L2-net: Deep learning of discriminative patch descriptor in euclidean
  space.
\newblock In: CVPR. (2017)

\bibitem{geodesc}
Luo, Z., Shen, T., Zhou, L., Zhu, S., Zhang, R., Yao, Y., Fang, T., Quan, L.:
\newblock Geodesc: Learning local descriptors by integrating geometry
  constraints.
\newblock In: ECCV. (2018)

\bibitem{caps}
Wang, Q., Zhou, X., Hariharan, B., Snavely, N.:
\newblock Learning feature descriptors using camera pose supervision.
\newblock In: ECCV. (2020)

\bibitem{sarlin2020superglue}
Sarlin, P.E., DeTone, D., Malisiewicz, T., Rabinovich, A.:
\newblock Superglue: Learning feature matching with graph neural networks.
\newblock In: CVPR. (2020)

\bibitem{chen2021sgm}
Chen, H., Luo, Z., Zhang, J., Zhou, L., Bai, X., Hu, Z., Tai, C.L., Quan, L.:
\newblock Learning to match features with seeded graph matching network.
\newblock In: ICCV. (2021)

\bibitem{zhang2019oanet}
Zhang, J., Sun, D., Luo, Z., Yao, A., Zhou, L., Shen, T., Chen, Y., Quan, L.,
  Liao, H.:
\newblock Learning two-view correspondences and geometry using order-aware
  network.
\newblock In: ICCV. (2019)

\bibitem{pointcn}
Yi*, K.M., Trulls*, E., Ono, Y., Lepetit, V., Salzmann, M., Fua, P.:
\newblock Learning to find good correspondences.
\newblock In: CVPR. (2018)

\bibitem{sun2020acne}
Sun, W., Jiang, W., Tagliasacchi, A., Trulls, E., Yi, K.M.:
\newblock Attentive context normalization for robust permutation-equivariant
  learning.
\newblock In: CVPR. (2020)

\bibitem{adalam}
Cavalli, L., Larsson, V., Oswald, M.R., Sattler, T., Pollefeys, M.:
\newblock Handcrafted outlier detection revisited.
\newblock In: ECCV. (2020)

\bibitem{Bian2020gms}
Bian, J., Lin, W.Y., Liu, Y., Zhang, L., Yeung, S.K., Cheng, M.M., Reid, I.:
\newblock {GMS}: Grid-based motion statistics for fast, ultra-robust feature
  correspondence.
\newblock IJCV (2020)

\bibitem{GOCor}
Truong, P., Danelljan, M., Gool, L., Timofte, R.:
\newblock Gocor: Bringing globally optimized correspondence volumes into your
  neural network.
\newblock In: NeurIPS. (2020)

\bibitem{flownet2}
Ilg, E., Mayer, N., Saikia, T., Keuper, M., Dosovitskiy, A., Brox, T.:
\newblock Flownet 2.0: Evolution of optical flow estimation with deep networks.
\newblock In: CVPR. (2017)

\bibitem{raft}
Teed, Z., Deng, J.:
\newblock Raft: Recurrent all-pairs field transforms for optical flow.
\newblock In: ECCV. (2020)

\bibitem{flownet}
Fischer, P., Dosovitskiy, A., Ilg, E., Häusser, P., Hazırbaş, C., Golkov,
  V., van~der Smagt, P., Cremers, D., Brox, T.:
\newblock Flownet: Learning optical flow with convolutional networks.
\newblock In: ICCV. (2015)

\bibitem{yin2018geonet}
Yin, Z., Shi, J.:
\newblock Geonet: Unsupervised learning of dense depth, optical flow and camera
  pose.
\newblock In: CVPR. (2018)

\bibitem{zhou2020kfnet}
Zhou, L., Luo, Z., Shen, T., Zhang, J., Zhen, M., Yao, Y., Fang, T., Quan, L.:
\newblock Kfnet: Learning temporal camera relocalization using kalman
  filtering.
\newblock In: CVPR. (2020)

\bibitem{prob1}
Gast, J., Roth, S.:
\newblock Lightweight probabilistic deep networks.
\newblock In: CVPR. (2018)

\bibitem{prob2}
Ilg, E., iek, z., Galesso, S., Klein, A., Makansi, O., Hutter, F., Brox, T.:
\newblock Uncertainty estimates and multi-hypotheses networks for optical flow.
\newblock In: ECCV. (2018)

\bibitem{prob3}
Danelljan, M., Gool, L., Timofte, R.:
\newblock Probabilistic regression for visual tracking.
\newblock In: CVPR. (2020)

\bibitem{resnet}
He, K., Zhang, X., Ren, S., Sun, J.:
\newblock Deep residual learning for image recognition.
\newblock In: CVPR. (2016)

\bibitem{dai2017scannet}
Dai, A., Chang, A.X., Savva, M., Halber, M., Funkhouser, T., Nie{\ss}ner, M.:
\newblock Scannet: Richly-annotated 3d reconstructions of indoor scenes.
\newblock In: CVPR. (2017)

\bibitem{li2018megadepth}
Li, Z., Snavely, N.:
\newblock Megadepth: Learning single-view depth prediction from internet
  photos.
\newblock In: CVPR. (2018)

\bibitem{yfcc}
Thomee, B., Shamma, D.A., Friedland, G., Elizalde, B., Ni, K., Poland, D.,
  Borth, D., Li, L.J.:
\newblock {YFCC100M}: The new data in multimedia research.
\newblock Communications of the ACM (2016)

\bibitem{pdcnet+}
Truong, P., Danelljan, M., Timofte, R., Van~Gool, L.:
\newblock {PDC-Net+}: Enhanced probabilistic dense correspondence network.
\newblock Preprint (2021)

\bibitem{wang2022matchformer}
Wang, Q., Zhang, J., Yang, K., Peng, K., Stiefelhagen, R.:
\newblock Matchformer: Interleaving attention in transformers for feature
  matching.
\newblock Preprint (2022)

\bibitem{dkm}
Edstedt, J., Wadenbäck, M., Felsberg, M.:
\newblock Deep kernelized dense geometric matching.
\newblock Preprint (2022)

\bibitem{taira2018inloc}
Taira, H., Okutomi, M., Sattler, T., Cimpoi, M., Pollefeys, M., Sivic, J.,
  Pajdla, T., Torii, A.:
\newblock Inloc: Indoor visual localization with dense matching and view
  synthesis.
\newblock In: CVPR. (2018)

\bibitem{zhang2021reference}
Zhang, Z., Sattler, T., Scaramuzza, D.:
\newblock Reference pose generation for long-term visual localization via
  learned features and view synthesis.
\newblock IJCV (2021)

\bibitem{Sattler2018CVPR}
Sattler, T., Maddern, W., Toft, C., Torii, A., Hammarstrand, L., Stenborg, E.,
  Safari, D., Okutomi, M., Pollefeys, M., Sivic, J., Kahl, F., Pajdla, T.:
\newblock {Benchmarking 6DOF Outdoor Visual Localization in Changing
  Conditions}.
\newblock In: CVPR. (2018)

\bibitem{toft2020long}
Toft, C., Maddern, W., Torii, A., Hammarstrand, L., Stenborg, E., Safari, D.,
  Okutomi, M., Pollefeys, M., Sivic, J., Pajdla, T.,  et~al.:
\newblock Long-term visual localization revisited.
\newblock TPAMI (2020)

\bibitem{sarlin2019coarse}
Sarlin, P.E., Cadena, C., Siegwart, R., Dymczyk, M.:
\newblock From coarse to fine: Robust hierarchical localization at large scale.
\newblock In: CVPR. (2019)

\bibitem{4th_imc}
Lashkov, I.:
\newblock 4th solution of imc 2022.
\newblock \url{
  https://www.kaggle.com/competitions/image-matching-challenge-2022/discussion/328805}

\end{thebibliography}

\end{document}